\documentclass[10pt,twocolumn,letterpaper]{article}

\usepackage[preprint,algorithms]{wacv}      

\usepackage{graphicx}
\usepackage{amsmath}
\usepackage{amssymb}
\usepackage{booktabs}
\usepackage[framemethod=tikz]{mdframed}
\usepackage{amsfonts}
\usepackage{lipsum}
\usepackage{diagbox}
\usepackage{tabularx}
\usepackage{array}
\usepackage{ragged2e}
\usepackage{multirow}
\usepackage{colortbl}
\usepackage{rotating}
\usepackage{booktabs}
\usepackage{caption}
\usepackage{lscape}
\usepackage{pdflscape}
\usepackage{footnote}
\usepackage{nomencl}
\usepackage{float}
\usepackage{siunitx}
\usepackage{rotating}

\usepackage[export]{adjustbox}
\usepackage[normalem]{ulem}

\useunder{\uline}{\ul}{}
\DeclareMathOperator*{\argmax}{\arg\!\max}

\usepackage[pagebackref,breaklinks,colorlinks]{hyperref}
\usepackage{orcidlink}
\usepackage{xcolor, colortbl}
\usepackage{arydshln} 
\usepackage[capitalize]{cleveref}
\crefname{section}{Sec.}{Secs.}
\Crefname{section}{Section}{Sections}
\Crefname{table}{Table}{Tables}
\crefname{table}{Tab.}{Tabs.}

\def\wacvPaperID{393} 
\def\confName{WACV}
\def\confYear{2025}

\begin{document}

\title{Visual Robustness Benchmark for Visual Question Answering (VQA)}

\author{Md Farhan Ishmam\thanks{Equal Contribution}~~\orcidlink{0009-0004-3725-0342}, Ishmam Tashdeed\footnotemark[1]~~\orcidlink{0009-0000-0114-6421}, Talukder Asir Saadat\footnotemark[1]~~\orcidlink{https://orcid.org/0009-0005-2495-6474}, \\
Md Hamjajul Ashmafee, Abu Raihan Mostofa Kamal~\orcidlink{0000-0001-9529-5208}, Md. Azam Hossain~\orcidlink{0000-0002-4315-5243}\\
\small Department of Computer Science and Engineering, Islamic University of Technology \\
{\tt\small farhanishmam, ishmamtashdeed, asirsaadat, ashmafee, azam, raihan.kamal@iut-dhaka.edu}
}

\maketitle

\begin{abstract}
   Can Visual Question Answering (VQA) systems perform just as well when deployed in the real world? Or are they susceptible to realistic corruption effects e.g. image blur, which can be detrimental in sensitive applications, such as medical VQA? While linguistic or textual robustness has been thoroughly explored in the VQA literature, there has yet to be any significant work on the visual robustness of VQA models. We propose the first large-scale benchmark comprising 213,000 augmented images, challenging the visual robustness of multiple VQA models, including multimodal large language models in zero-shot settings, and assessing the strength of realistic visual corruptions. Additionally, we have designed several robustness evaluation metrics that can be aggregated into a unified metric and tailored to fit a variety of use cases. Our experiments reveal important insights into the relationships between model size, performance, and robustness with the visual corruptions. Our benchmark highlights the need for a balanced approach in model development that considers model performance without compromising the robustness.
   

\end{abstract}

\section{Introduction}
\label{sec:Intro}

   \begin{figure*}[ht]
        \centering
            \includegraphics[width=0.98\textwidth]{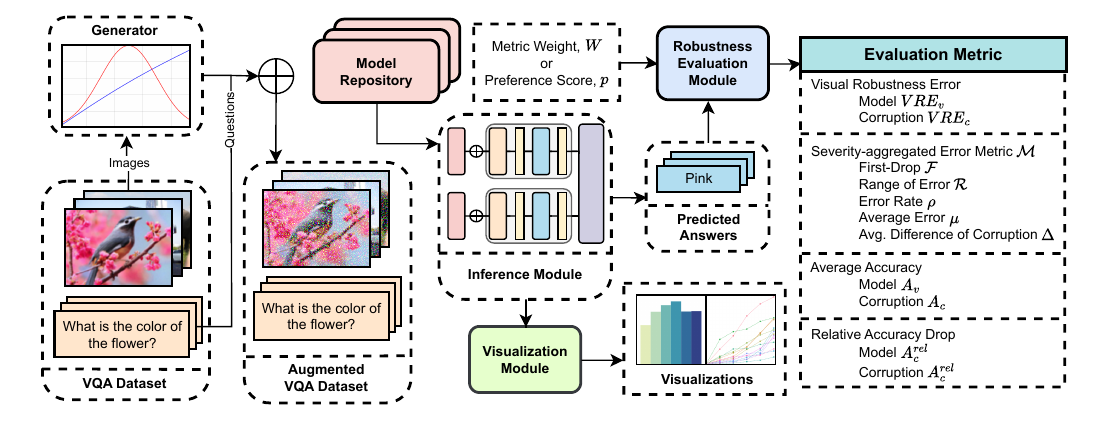}
        \caption{Architecture of the Visual Robustness Framework and its components -- \textbf{Model Repository:} Hosts multiple VQA models for inference, \textbf{Generator:} Applies the corruption functions to the VQA dataset and generate multiple augmented datasets, \textbf{Inference Module:} Inferences on the augmented datasets using models from the model repository, \textbf{Robustness Evaluation Module:} Evaluates the results of the inference module by computing different robustness metrics introduced in this work, \textbf{Visualization Module:} Produces visualizations based on the predicted answers. The VQA datasets, models, and corruptions are the input to the framework while the VRE scores, accuracy scores, and visualizations will be produced as the output.}
        \label{fig:archi}
    \end{figure*} 

    Visual Question Answering (VQA) involves answering any question based on a visual input provided as the context \cite{antol2015vqa}, extending the problem of contextual question answering \cite{choi2018quac}. In recent years, VQA has been widely adopted in several emerging applications \cite{barra2021visual, ishmam2024image}, such as assisting the visually impaired \cite{gurari2018vizwiz}, medical visual questions answering \cite{abacha2019vqaMed}, and visual chat-bots \cite{das2017visualDialog}. As VQA requires a precise understanding of both visual and textual modalities and their interrelationship, affecting any one or both modalities can lead to unsatisfactory performance \cite{agrawal2018don, ramakrishnan2018overcoming}.

    The deep learning architectures employed in VQA have been prone to adversarial attacks, corruption effects, and noise \cite{sharma2018attend, huang2019novel, jimenez2022carets}. These effects can be textual \eg adding grammatical errors to the questions or visual \eg changing the brightness of the image. Robustness is the ability of the model to resist such adversarial effects, which has been the subject of interest in several vision \cite{szegedy2013intriguing, hendrik2017universal, moosavi2017universal} and language \cite{jia2017adversarial, zhang2020adversarial, morris2020textattack} tasks. Robustness is crucial in sensitive VQA applications and assessing model performance during deployment. In the VQA domain, several works focused on textual robustness \cite{huang2019novel, rosenberg2021vqa, jimenez2022carets}, with only a few addressing visual robustness \cite{gupta2022grit} with limited samples and metrics. 

    For evaluation, VQA uses accuracy \cite{antol2015vqa} to estimate the performance in ideal conditions but not for robustness evaluation. Furthermore, robustness evaluation is not straightforward, requiring quantification of multiple factors or a multi-faceted metric that can vary based on the use case \cite{mcphail2018robustness, brendel2019accurate,guo2023comprehensive}. \Eg some systems may evaluate robustness based on the resilience to corruption effects at lower magnitudes while others may prioritize the robustness based on maintaining performance over varying intensities.

    Addressing the above issues, we propose a framework, depicted in \cref{fig:archi}, to evaluate visual robustness in the domain of VQA. Our contributions can be summarized as: 

    \begin{enumerate}
        \item We establish the first large-scale visual robustness evaluation to evaluate VQA models' robustness to real-world visual corruptions and access the strength of the corruptions. Our framework comprises 213,000 augmented images from 3,000 unique images and their corresponding 16,000 QA pairs.

        \item We propose $5$ novel visual robustness evaluation metrics aggregated into a unified metric called Visual Robustness Error (VRE) for model robustness and corruption strength evaluation. VRE can be customized using preference values based on the specific use case.
        
        \item We designed a modular and extensible robustness evaluation framework, rigorously tested $6$ VQA models, including $2$ Multimodal Large Language Models (MLLMs), assessed the strength of $14$ visual corruption effects, and highlighted relationships between model size, performance, and robustness with the visual corruptions.
    \end{enumerate}

\section{Related Work}
\label{sec:RelatedWork}
    \noindent \textbf{Visual Question Answering (VQA).} VQA requires an in-depth understanding of both visual and textual modalities \cite{goyal2017making}. Over the years, several works \cite{fukui2016multimodal, yang2016stacked, ben2017mutan, anderson2018bottom} constantly improved the accuracy of VQA models. Transformers \cite{vaswani2017attention} and Vision Transformers \cite{dosovitskiy2020image} ushered in a new wave of transformer-based methods for VQA \cite{ lu2019vilbert, li2019visualbert, kim2021vilt, wang2021vlmo}. These methods relied on pre-training for language processing \cite{lu2019vilbert}, patched image processing using self-attention \cite{kim2021vilt}, and unified vision-language pre-training (VLP) \cite{zhou2020unified, chen2023vlp}. Currently, the Vision-Language models leverage VLP in a unified transformer architecture \cite{zhou2020unified} which is fine-tuned on downstream tasks, such as VQA \cite{tan2019lxmert, wang2021vlmo}.
    \\ 
    
   \noindent \textbf{Robustness in VQA.} The textual robustness of VQA models has been explored using several strategies, such as semantically similar questions \cite{huang2019novel}, adversarial attacks by humans \cite{li2021adversarial}, counterfactually augmenting the data \cite{chen2020counterfactual, rosenberg2021vqa}, and rephrasing the question \cite{jimenez2022carets}. On the contrary, works on visual robustness were limited to blurring, cropping, and masking irrelevant parts of the image \cite{jimenez2022carets}, and using visual corruptions but limited to $3$ samples only \cite{gupta2022grit}.
   \\

    \noindent \textbf{Robustness in Computer Vision (CV).} 
    The robustness of image classification models was introduced by adding perturbations to neural networks \cite{szegedy2013intriguing}. Later works focused on benchmarking robustness using realistic visual corruptions and perturbations for several CV tasks, such as image classification \cite{hendrycks2019benchmarking}, semantic segmentation \cite{kamann2021benchmarking}, object detection \cite{dong2023benchmarking,liu2024benchmarking}, and autonomous driving \cite{michaelis2019benchmarking, dong2023benchmarking}. Generalized robustness benchmarks for CV were also proposed \cite{mu2019mnist, gupta2022grit}.
    \\

        \noindent \textbf{Robustness Evaluation Metrics.}
        The error-based metrics -- mean Corruption Error (mCE) and relative mCE were introduced to measure robustness for common corruptions for image classification \cite{hendrycks2019benchmarking}. The metrics are derived from top-1 error and may be unsuitable for VQA classification which uses VQA accuracy \cite{antol2015vqa}. Furthermore, both metrics depend on a reference model in that domain to measure the robustness. Self-consistency and comprehensive accuracy are consistency-based metrics that can be used as robustness measures \cite{jimenez2022carets}. The $R_{\text{score}}$ is also a popular robustness metric in VQA but for the textual modality only \cite{huang2019novel}.

\section{Methodology}
\label{sec:Methodology}

    \subsection{Visual Question Answering (VQA)}
    \label{sec:visualQuestionAnswering}

    VQA is defined as the classification problem of selecting an answer from a set of predefined answer classes, given a visual input and a textual question. Although multiple variations of VQA are observed across several modalities \cite{ishmam2024image}, we limit the visual and textual inputs to a single image and question respectively. 
    The trained VQA classifier $\mathit{V}: (I, Q) \xrightarrow{} Y$ with parameters $\theta$ selecting an answer $y$ for the image-question pair $(i,q)$, can be formulated as:
    \begin{equation}
    \label{eq:VQAclassifier}
    V(i,q) = \underset{y_i \in Y}{\argmax{}}P(y_i|i,q; \theta)
    \end{equation}
    
    \subsection{Visual Robustness in VQA}
     Visual corruption is defined as the degradation of an image due to various factors, such as lighting effects, weather conditions, and compression artifacts. Inspired by \cite{hendrycks2019benchmarking}, our work primarily focuses on the prevalent forms of degradation commonly observed in real-world scenarios \eg noise, blur, pixelation, and weather effects, differing significantly from adversarial attacks \cite{szegedy2013intriguing, jia2017adversarial} described later in this subsection but not covered in our work.
     
     Given, a set of visual corruption functions $\mathcal{C}$, each associated with a probability of occurring in real life $\mathbb{P}_{\mathcal{C}}(c)$, where $c \in \mathcal{C}$. We derive the uncorrupted test accuracy on distribution $\mathcal{D}$ as $\mathbb{P}_{(i,q,y)\sim \mathcal{D}}(\mathit{V}(i,q)=y)$. The visual corruption robustness of our VQA classifier can be formulated as, 
    $
         \mathbb{E}_{c\sim\mathcal{C}}[P_{(i,q,y)\sim \mathcal{D}}(\mathit{V}(c(i),q)=y)]
    $, corresponding to the average case performance. 
    On the contrary, adversarial attacks correspond to the worst-case performance on the classifier's response to small and specific changes made to the input data, and such changes are tailored to that specific classifier instead of affecting them universally. Adversarial robustness is formulated as, 
    $
    \operatorname{min}_{||\delta||_{p}<\mathit{\epsilon}}[\mathbb{P}_{(i,q,y)\sim \mathcal{D}}(\mathit{V}(i+\delta_i,q+\delta_q)=y)]
    $, where $\mathit{\epsilon}$ is a small budget/increment. Since we are concerned about the performance of our models on real-world corruption effects, robustness is evaluated as the average-case performance instead of the worst-case performance. 

    \begin{figure}[ht]
      \centering
        \includegraphics[width=0.45\textwidth]{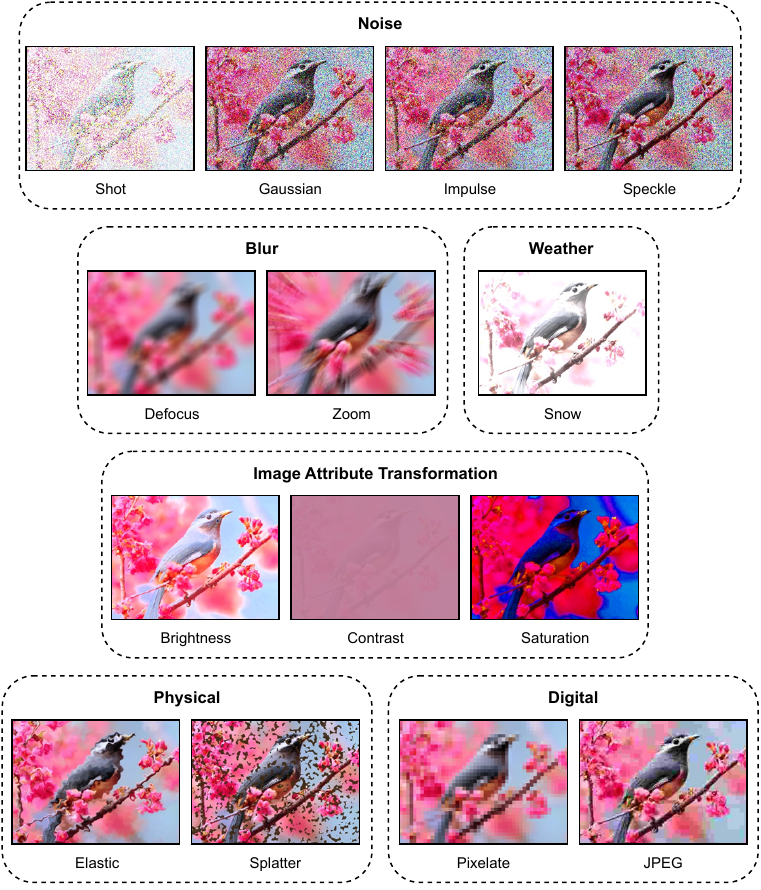}
    \caption{Effects of the $14$ visual corruption functions at the highest severity level $5$ introduced in our work. The functions are categorized into $6$ corruption classes.}
    \label{fig:vis}
    \end{figure}


    \subsection{Visual Corruption Functions}
    Our experiments were conducted using $14$ visual corruption functions to create a diverse set of artificially corrupted image datasets. The functions have severity levels from $1$ to $5$ and an additional base level $0$, denoting uncorrupted or clean images. The properties of each severity level for a particular corruption function depend on the parametric values of that function. The functions and the severity level parameter values are chosen based on replicating corruption effects resembling realistic conditions from ImageNet-C \cite{hendrycks2019benchmarking} and Imgaug \cite{imgaug}. The categorization and definitions of the corruption functions are provided in \cref{app:visualCorruptionFunctions}.

\subsection{Visual Robustness Evaluation Suite}

    Our evaluation suite requires a set of corruption functions at multiple severity levels, a set of VQA models, and one or more datasets. Additionally, the components of the evaluation suite should be \emph{modular} \ie the components can be used interchangeably in different variations of the framework, and \emph{extensible} \ie additional components can enhance the functionality of the framework. Considering these design aspects, the Visual Robustness Evaluation Suite has been developed and its architecture is depicted in \cref{fig:archi}. The primary purpose of the suite is to evaluate the visual robustness of VQA models and the strength of corruption functions. The framework's inherent extensibility implies that the evaluation is not limited to the task of VQA only but to any vision-language task.

\section{Accuracy-based Evaluation Metrics}
For ease of understanding the aforementioned metrics, a nomenclature has been provided in \cref{app:evalMetrics}.

    \subsection{Accuracy}
        \label{sec:accuracy}
    Following \cref{sec:visualQuestionAnswering}, as VQA is defined as a multiclass classification problem, misclassification accuracy or top-1 accuracy is the primary choice for model evaluation. Here, the answer class predicted with the highest probability \textit{exactly} matches the ground truth answer. Due to variations of answers to a single question, \cite{antol2015vqa} derived VQA accuracy for evaluating open-ended tasks. VQA accuracy is the standard evaluation metric for the VQAv2 dataset \cite{goyal2017making} and can be mathematically expressed as:
    
    \begin{equation}
    \label{eq:VQAaccuracy}
      A_{v,c,l} =  \frac{1}{N_Q}\sum_{q \in \mathcal{Q}}
        \min\left(  
            \frac{
                \sum_{a \in \mathcal{A}_q} \mathbb{I}[\hat{a}_q = a_q]
            }{3}
        ,1\right)  
    \end{equation}
     where, $N_Q$ is the number of questions in the dataset, $q$ is a particular question from the question set $\mathcal{Q}$, $a_q$ is a particular human annotated answer to $q$ from the corresponding answer set $\mathcal{A}_q$, $\mathbb{I}[\cdot]$ is the indicator function, and $\hat{a}$ is the predicted answer for $q$. 
    
     \subsection{Average Accuracy}
     \label{sec:averageAccuracy}
     The severity-aggregated average accuracy for a VQA model $v \in \mathcal{V}$ and corruption $c \in \mathcal{C}$ can be computed as:
    \begin{equation}
        A_{v,c} = \frac{1}{L} \underset{l\in \mathcal{L}}{\sum} A_{v,c,l} \label{eq:sevAggaverageAccuracy} 
    \end{equation}
    Furthermore, we compute model-wise and corruption-wise average accuracy:
    \begin{equation}  
    A_v = \frac{1}{C}\underset{c\in \mathcal{C}}{\sum} A_{v, c} \label{eq:modelAggaverageAccuracy}
    \end{equation}
    \begin{equation} 
    A_c = \frac{1}{V}\underset{v\in \mathcal{V}}{\sum} A_{v, c} \label{eq:corrAggaverageAccuracy}
    \end{equation}

    Average accuracy can be a simple indicator of model performance but may not be suitable for robustness evaluation in some scenarios (Appendix \ref{app:robustnessScenarios}).
  
\subsection{Relative Accuracy Drop}
    Similar to relative error, a well-established metric in measurement, we derive the relative accuracy drop for severity-aggregated average accuracy as follows:
    \begin{equation}
    \label{eq:relAccuracy}
        A^{rel}_{v,c} = \frac{A_{v,0} - A_{v,c}}{A_{v,0}}
    \end{equation}
    Relative accuracy drop for models can be formulated as:
    \begin{equation}
    \label{eq:relAccuracyModel}
        A^{rel}_v = \frac{A_{v,0} - A_v}{A_{v,0}}
    \end{equation}
    
    Since visual corruption effects lack a base accuracy, we combine the differences between the model's base accuracy and severity-aggregated average accuracy to formulate relative accuracy for corruptions as follows:
    
     \begin{equation}
         \label{eq:relAccuracyCorr}
         A^{rel}_c =\frac{\underset{v \in \mathcal{V}}{\sum} \left(A_{v,0} - A_{v,c}\right)}{\underset{v \in \mathcal{V}}{\sum} A_{v,0}}
     \end{equation}

     The relative accuracy drop decouples the base performance from the overall evaluation, instead aggregating the deviation from the base performance to provide a fairer evaluation.

\section{Error-based Evaluation Metrics}

    To evaluate visual robustness, the top-1 error is used as the standard error metric \cite{hendrycks2019benchmarking} which can be derived from top-1 accuracy. However, for VQA, we use the error equivalent of the VQA accuracy from \cref{eq:VQAaccuracy}, as follows:
    \begin{equation} 
    \label{eq:error}
        E_{v,c,l} = 1 - A_{v, c, l}
    \end{equation}

    \subsection{Severity-aggregated Metrics}
    \noindent\textbf{First-Drop.}
    The metric quantifies the error when a model is introduced to low levels of corruption. We define it as the relative change in error when exposed to the minimal degree of corruption \ie the difference between the level-1 and level-0 (clean) error relative to the level-0 error.
     \begin{equation} 
        \mathcal{F}_{v,c} = \frac{E_{v,c,1} - E_{v,c,0}}{E_{v,c,0}}
    \end{equation}
  
    \noindent\textbf{Range of Error.}
    The range of error reflects the degree to which a model's performance degrades across varying severity levels, determined by the relative difference between the maximum and minimum error values \ie the range.
    \begin{equation} 
    \label{eq:errorRange}
        \mathcal{R}_{v,c} = \frac{\underset{l\in\mathcal{L}}{\operatorname{max}\ }E_{v,c,l} - \underset{l\in\mathcal{L}}{\operatorname{min}\ }E_{v,c,l}}{\underset{l\in\mathcal{L}}{\operatorname{min}\ }E_{v,c,l}}
    \end{equation} 

    \noindent\textbf{Error Rate.}
    Error rate signifies the rate of change for error values, calculated as the slope of the best-fit line through the error values across severity levels. 

    \begin{equation} 
        \rho_{v,c} = \frac{
            L\underset{l\in \mathcal{L}}{\sum}
            \left(l\cdot E_{v,c,l}\right)-
            \left(\underset{l\in \mathcal{L}}{\sum}l\right)\cdot
            \left(\underset{l\in \mathcal{L}}{\sum}E_{v,c,l}\right)}
            {L\underset{l\in \mathcal{L}}{\sum}l^2
        - \left(\underset{l\in \mathcal{L}}{\sum}l\right)^2}
 \end{equation}
    
    \noindent\textbf{Average Error.} A simple metric that uniformly encapsulates all severity levels. Defined as the error-equivalent of average accuracy from \cref{eq:sevAggaverageAccuracy}.
    \begin{equation} 
    \label{eq:averageError}
        \mu_{v,c} = \frac{1}{L}\underset{l\in \mathcal{L}}{\sum}E_{v,c,l}
    \end{equation}

    \noindent\textbf{Average Difference of Corruption Error.}
    A more nuanced metric, inspired by \cite{hendrycks2019benchmarking}, aggregates the error differences with the uncorrupted/clean image, thereby capturing the average deviation from the uncorrupted performance.
    \begin{equation} 
        \Delta_{v,c} = \frac{1}{L}\underset{l\in \mathcal{L}'}{\sum}\left(E_{v,c,l}-E_{v,c,0}\right)
    \end{equation}
  
    
    \subsection{Metric Scaling}
    The error values produced by each severity-aggregated metric $\mathcal{M}_{v,c}$ have different ranges, evident in the Appendix \cref{tab:metric}. The variations in ranges cause difficulty in metric aggregation and VRE computation. Hence, we scale the metrics to $[0,1]$ using min-max scaling on the non-scaled generalized metric $\mathcal{M'} \in \mathbb{M}$, preserving the shape of the distribution. The scaled generalized metric can be formulated as:
    \begin{equation}
        \mathcal{M}_{v,c} = \frac{\mathcal{M'}_{v,c} - \underset{v,c}{\min}\ \mathcal{M'}_{v,c}}{\underset{v,c}{\max}\ \mathcal{M'}_{v,c} - \underset{v,c}{\min}\ \mathcal{M'}_{v,c}}
    \end{equation}

\definecolor{myred}{rgb}{1.0, 0.9, 0.9}
\definecolor{mygreen}{rgb}{0.9, 1.0, 0.9}  

\begin{table*}[ht]
\centering

\resizebox{\textwidth}{!}{%
\begin{tabular}{cccccccccccccccccc}
\toprule
\multirow{2}{*}{\textbf{Model}} & \multicolumn{4}{c}{\textbf{Noise}}             & \multicolumn{2}{c}{\textbf{Blur}} & \textbf{Weather} & \multicolumn{3}{c}{\textbf{Image Attribute}} & \multicolumn{2}{c}{\textbf{Physical}} & \multicolumn{2}{c}{\textbf{Digital}}       & \multicolumn{1}{c}{\multirow{2}{*}{$A_{v}$ \textcolor{green}{$\uparrow$}}} & \multicolumn{1}{c}{\multirow{2}{*}{$A_{v,0}$ \textcolor{green}{$\uparrow$}}} & \multirow{2}{*}{$A^{rel}_v (\%)$ \textcolor{red}{$\downarrow$}} \\ \cmidrule(lr){2-5} \cmidrule(lr){6-7} \cmidrule(lr){8-8} \cmidrule(lr){9-11} \cmidrule(lr){12-13} \cmidrule(lr){14-15}
                       & Shot           & Gaus  & Imp   & Spec  & Defoc       & Zoom        & Snow    & Brig             & Cont    & Sat     & Elas          & Spl           & Pix   & \multicolumn{1}{c}{JPEG}  & \multicolumn{1}{c}{}                      & \multicolumn{1}{c}{}                      &                   \\ \midrule
ViLT                   & 0.576          & 0.664 & 0.659 & 0.685 & 0.672       & 0.563       & 0.603   & 0.700            & 0.639   & 0.680   & 0.688         & 0.664         & 0.706 & \multicolumn{1}{c}{0.705} & \multicolumn{1}{c}{0.657}                 & \multicolumn{1}{c}{0.713}                 & \cellcolor{mygreen}\textbf{07.76}    \\
BLIP                   & 0.627          & 0.716 & 0.708 & 0.733 & 0.707       & 0.626       & 0.687   & 0.756            & 0.716   & 0.731   & 0.730         & 0.728         & 0.736 & \multicolumn{1}{c}{0.739} & 0.710           & 0.782           & 09.29             \\
VLE                    & 0.567          & 0.673 & 0.669 & 0.701 & 0.674       & 0.602       & 0.698   & 0.759            & 0.718   & 0.740   & 0.715         & 0.736         & 0.729 & \multicolumn{1}{c}{0.715} & \multicolumn{1}{c}{0.693}                 & \multicolumn{1}{c}{0.771}                 & 10.17             \\ \hdashline[3pt/1.25pt]
PNP                    & 0.418          & 0.449 & 0.447 & 0.454 & 0.447       & 0.424       & 0.436   & 0.460            & 0.444   & 0.445   & 0.454         & 0.449         & 0.458 & \multicolumn{1}{c}{0.459} & \cellcolor{myred}\textbf{0.446}        & \cellcolor{myred}\textbf{0.648}        & \cellcolor{myred}\textbf{31.19}       \\ \hdashline[3pt/1.25pt]

LLaVA-7B & 0.626 & 0.716  & 0.698 & 0.722 & 0.701  & 0.651 &  0.690 &  0.769 & 0.726 & 0.741 & 0.742 & 0.729 & 0.768 & 0.770 & 0.718 & 0.783 & 08.35 \\ 
LLaVA-13B & 0.635 & 0.718& 0.699  & 0.731& 0.701& 0.653& 0.700  & 0.782& 0.738& 0.748& 0.751& 0.736& 0.778& 0.782 & \cellcolor{mygreen}{\textbf{0.725}} & \cellcolor{mygreen}{\textbf{0.798}} & 09.21\\ \midrule

$A_c$ \textcolor{green}{$\uparrow$}                  & \cellcolor{myred} \textbf{0.575} & 0.656 & 0.647 & 0.671  & 0.650       & 0.586       & 0.636   & \cellcolor{mygreen} \textbf{0.704}      & 0.663   & 0.681   & 0.680         & 0.674         & 0.696 & 0.695                      &                                            &                                            &                   \\ \cmidrule{1-15}

$A^{rel}_c (\%)$ \textcolor{red}{$\downarrow$}                       & \cellcolor{myred} \textbf{23.61} & 12.87& 14.08&  10.90 & 13.61& 22.12& 15.66 &  \cellcolor{mygreen}\textbf{06.62}& 12.01&09.73&  09.75& 10.62&  07.67&  07.80   \\ \cmidrule{1-15}     
\end{tabular}%

}
\caption{Severity-aggregated average accuracy, base accuracy, average accuracy, and relative accuracy drop values for different models and corruptions. \colorbox{mygreen}{\textbf{Green}} and \colorbox{myred}{\textbf{red}} indicate the maximum and minimum respectively for the base accuracy, average accuracy, or relative accuracy drop of that particular corruption or corruption effect. \textcolor{green}{$\uparrow$} indicates higher is better while \textcolor{red}{$\downarrow$} indicate the opposite.}

\label{tab:accVRETable}
\end{table*}

    \subsection{Corruption and Model-aggregated Metrics}
    The corruption-aggregated metrics average over corruption types $c \in \mathcal{C}$ while the model-aggregated metrics average over models $v \in \mathcal{V}$, both for the generalized metrics $\mathcal{M}_{v,c}$. The aggregated metrics measure an aspect of robustness for a corruption or model.
    \begin{equation}
    \label{eq:CorAggMetric}
        \mathcal{M}_v = 
        \frac{1}{C}\underset{c\in \mathcal{C}}{\sum}\mathcal{M}_{v,c}
    \end{equation}
    \begin{equation}
    \label{eq:modelAggMetric}
        \mathcal{M}_c = \frac{1}{V}\underset{v\in \mathcal{V}}{\sum}\mathcal{M}_{v,c}
    \end{equation}

    \subsection{Visual Robustness Error (VRE)}
     VRE for a particular model $v$ or visual corruption $c$ is defined as the weighted average of the corruption or model-aggregated metrics, mathematically represented as:
    \begin{equation}
    \label{eq:vreV}
        \operatorname{VRE}_v =  \underset{\mathcal{M}\in \mathbb{M}}{\sum}W_\mathcal{M}\mathcal{M}_v
         \text{  where  } \underset{\mathcal{M}\in \mathbb{M}}{\sum}W_\mathcal{M} = 1
    \end{equation}
    \begin{equation}
        \operatorname{VRE}_c =  \underset{\mathcal{M}\in \mathbb{M}}{\sum}W_\mathcal{M}\mathcal{M}_c
         \text{  where  } \underset{\mathcal{M}\in \mathbb{M}}{\sum}W_\mathcal{M} = 1
    \end{equation} 

By default, the weight vector is equally distributed among the five sub-metrics \ie $W = [0.2, 0.2, 0.2, 0.2, 0.2]$. The weight vector can be changed based on the evaluation requirement of the user and can be calculated from the preference scores, $p_{\mathcal{M}}$, assigned to each sub-metric by the user:
\begin{align}
    W_{\mathcal{M}} &= \underset{\mathcal{M}}{\operatorname{softmax}}(p_{\mathcal{M}}) \\
    &= \frac{e^{p_{\mathcal{M}}}}{\underset{m \in \mathbb{M}}{\sum} e^{p_m}}
\end{align}
The softmax function ensures non-linear scaling of the preference scores while satisfying $\sum W_\mathcal{M} = 1$. Trivially, for the default weight vector, the preference scores are equal.

\section{Experimental Setup}

\noindent \textbf{Computation Details.}
The benchmarking was performed on Nvidia V100 32 GB GPUs following the Hugging-Face implementation of the models \cite{wolf-etal-2020-transformers}. The pre-trained weights of the models were used in the default configuration.

\vspace{0.5cm}
\noindent \textbf{Datasets.}
Our framework evaluated VQA models on $3000$ images and $16000$ question-answer pairs sourced from the benchmark dataset VQAv2 \cite{goyal2017making}. The images were augmented using $14$ transformation functions at $5$ severity levels. Excluding the uncorrupted dataset, every model was tested on $14\times5=70$ augmented datasets resulting in a total of $70\times3000 = 210000$ augmented images.

\vspace{0.5cm}
\noindent \textbf{Models.}
We evaluated $3$ standard VQA models: ViLT \cite{kim2021vilt}, BLIP \cite{li2022blip}, VLE \cite{iflytekVLE}, $3$ zero-shot VQA (ZS-VQA) models -- PNP \cite{tiong2022plug}, and $2$ Multimodal Large Language Models (MLLMs) \cite{fu2023mme,achiam2023gpt} -- the 7B and 13B variants of LLaVA-1.5\cite{liu2024improved}. BLIP, ViLT, and VLE produce remarkably high performance in VQA benchmarks. All three models leverage the power of transformer architectures \cite{vaswani2017attention} for visual and textual processing. PNP employs a modular architecture, allowing greater flexibility and customization. Our implementation of the PNP model utilized UnifiedQA \cite{khashabi2022unifiedqa} as the question-answering module and BLIP \cite{li2022blip} for image-question matching and image captioning \cite{hossain2019comprehensive} modules. The LLaVA models utilize the corresponding 7B and 13B variants of the Vicuna LLM backbone \cite{zheng2023judging}. 



\section{Result Analysis}




\begin{figure}[ht]
  \centering
  \includegraphics[width=0.49\textwidth]{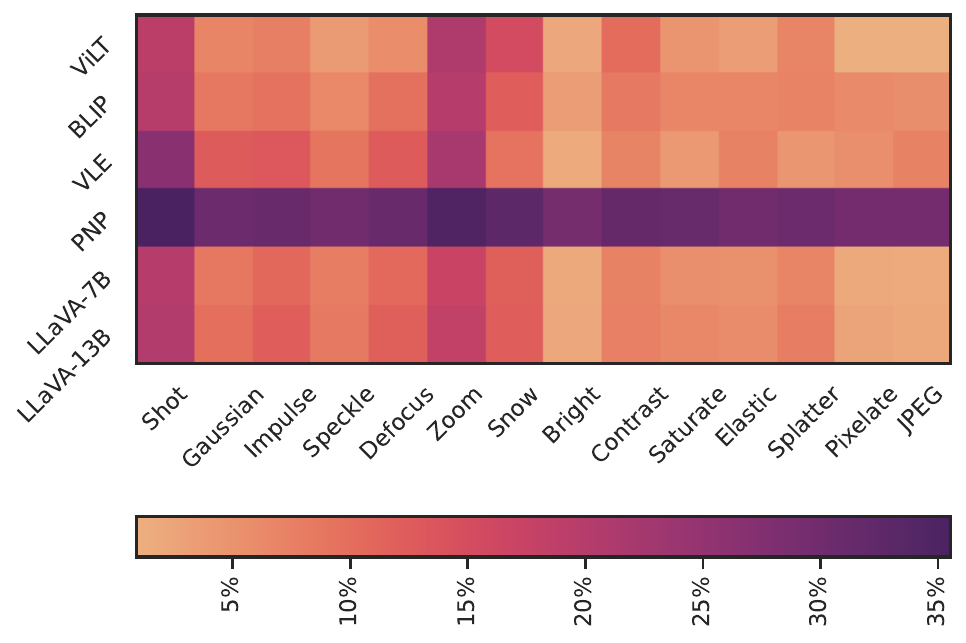}
  \caption{Visualization of relative accuracy drop $A^{rel}_{v,c}$ of the models and visual corruptions. Significantly higher drops are observed for PNP \cite{tiong2022plug}, shot noise, and zoom blur compared to other models and corruptions, while the brightness corruption induces a negligible accuracy drop for the standard VQA models.}
  \label{fig:heatmap}
\end{figure}

\begin{figure*}[ht]
  \centering
  \includegraphics[width = \textwidth]{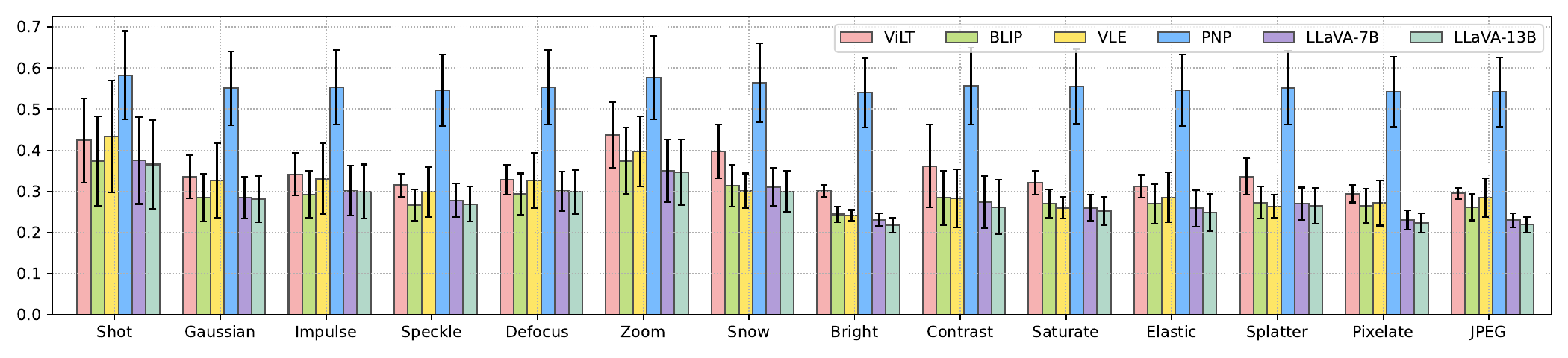}
  \caption{Average error $\mu_{v,c}$ of various models and visual corruption effects. The average error is the complement of average accuracy and can serve as a simple performance indicator for assessing model robustness.}
  \label{fig:errorBar}
\end{figure*}
    
\subsection{Effect of Visual Corruptions on Accuracy}
    Intuitively, introducing visual corruption should lead to a decrease in accuracy for all the models. Our findings in \cref{tab:accVRETable} confirm this, revealing a $7-10\%$ relative accuracy drop for the standard VQA models and a $31\%$ drop for the ZS-VQA model, PNP. We uncover a substantial performance gap between standard VQA models, including MLLMs, and ZS-VQA models, reinforced by the observations in \cref{fig:heatmap} for all types of visual corruptions. The accuracy drops among the standard VQA models and MLLMs are relatively similar.

    \subsection{Model Robustness}
\begin{figure*}[ht]
  \centering
  \includegraphics[width=0.85\textwidth]{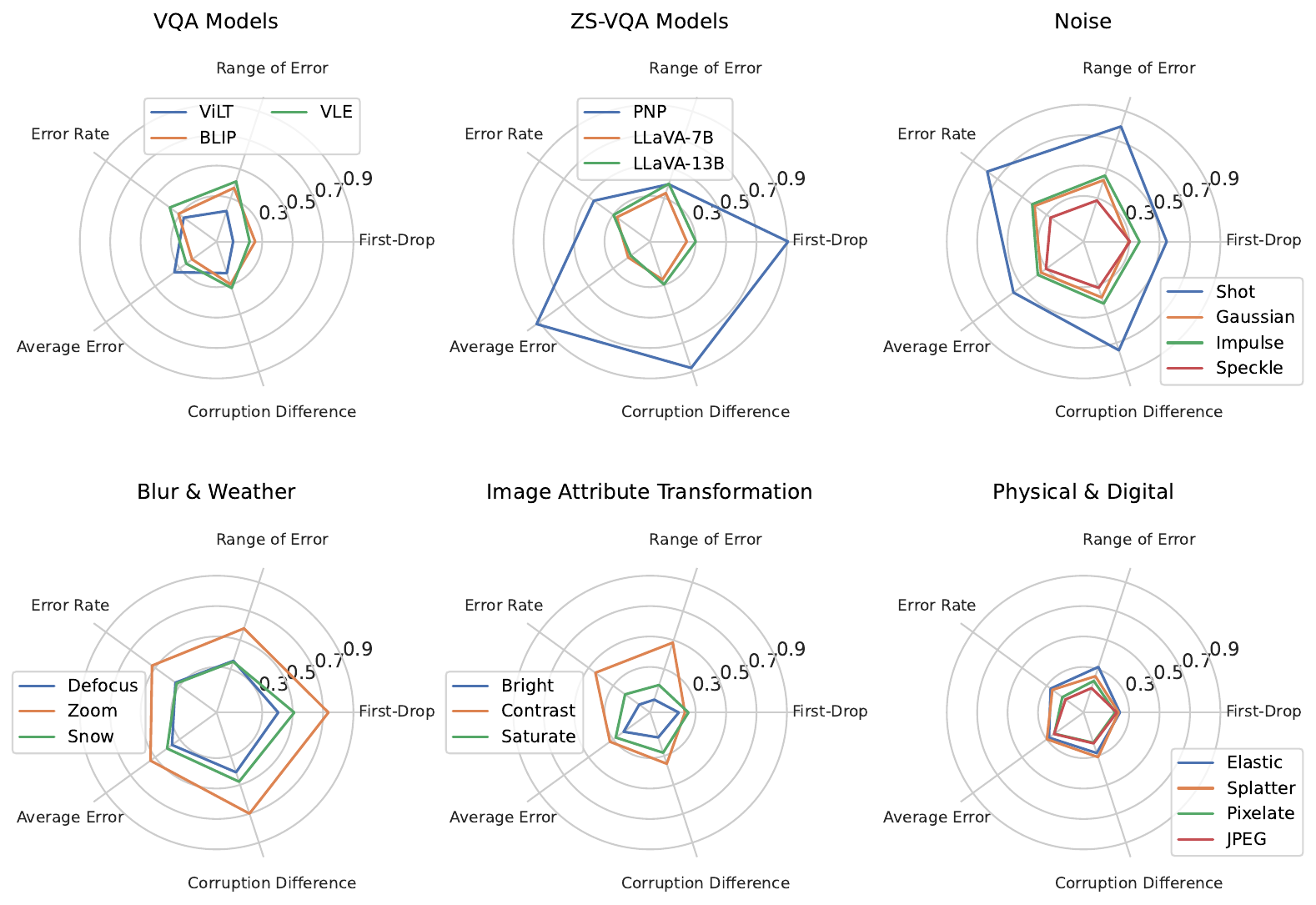}
  \caption{Composition of normalized metric values of the five severity-aggregated sub-metrics composing VRE for the models and different categories of visual corruptions.}
  \label{fig:metricComposition}
\end{figure*}

    \cref{fig:errorBar} illustrates the ZS-VQA model PNP producing significantly higher average error compared to other models. This is affirmed by \cref{fig:metricComposition}, where PNP has higher values in most robustness sub-metrics while encompassing a larger area. Based on the average error, ViLT appears to be the least robust standard VQA model but \cref{tab:metric,fig:metricComposition} confirms ViLT as the most robust model based on the other severity-aggregated metrics and the VRE metric. The robustness of BLIP, VLE, and LLaVA variants is close to ViLT based on the VRE metric.  

    \definecolor{mygrey}{gray}{0.93}

\begin{table}[ht]
\centering
\resizebox{0.49\textwidth}{!}{%
\begin{tabular}{cccccccccc}
\toprule
\multicolumn{2}{c}{\multirow{2}{*}{\begin{tabular}[c]{@{}c@{}}\textbf{Metrics}\end{tabular}}} & \multicolumn{2}{c}{\textbf{Accuracy Metrics}}                   & \multicolumn{6}{c}{\textbf{Non-normalized Error Metrics}}                                                    \\ 
\cmidrule(lr){3-4} \cmidrule(lr){5-10} 
\multicolumn{2}{c}{}                                                                             & $A_{v/c}$ \textcolor{green}{$\uparrow$}              & $A_{v/c}^{rel}$ \textcolor{red}{$\downarrow$}              & $\mathcal{F}$ \textcolor{red}{$\downarrow$}              & $\mathcal{R}$ \textcolor{red}{$\downarrow$}              & $\rho$ \textcolor{red}{$\downarrow$}              & $\mu$ \textcolor{red}{$\downarrow$}              & \multicolumn{1}{c}{$\Delta$ \textcolor{red}{$\downarrow$}}              & VRE \textcolor{red}{$\downarrow$}            \\ \midrule
\multirow{6}{*}{\rotatebox[origin=c]{90}{\textbf{Models}}}                                   & ViLT                                                   & 0.657          & \cellcolor{mygrey}\textbf{0.076} & \cellcolor{mygrey} \textbf{0.053} & \cellcolor{mygrey}\textbf{0.467} & \cellcolor{mygrey}\textbf{0.026} & 0.343          &  \cellcolor{mygrey} \textbf{0.066} & \cellcolor{mygrey}\textbf{0.230} \\
                                                     & BLIP                                        & 
 0.710 & 0.093                               & 0.160          & 0.712          & 0.029          & 0.290 & \multicolumn{1}{c}{0.087}          & 0.286          \\
                                                     & VLE                                                & 0.693          & \multicolumn{1}{c}{0.102}          & 0.133          & 0.781          & 0.034          & 0.307          & \multicolumn{1}{c}{0.094}          & 0.316          \\
                                                     & PNP                                          & 0.446          & \multicolumn{1}{c}{0.312}          & 0.644          & 0.751          & 0.040          & 0.554          & \multicolumn{1}{c}{0.243}          & 0.712          \\
                                                     & LLaVA-7B                                                 & 0.718          & \multicolumn{1}{c}{0.083}          & 0.151          & 0.657          & 0.026          & 0.282          &    0.078      &   0.258       \\
                                                     & LLaVA-13B                                                & \cellcolor{mygrey} \textbf{0.725}          & \multicolumn{1}{c}{0.092}          & 0.195         &    0.756       &  0.028         &   \cellcolor{mygrey} \textbf{0.275}        &   0.087    &      0.290    \\\midrule
\multirow{14}{*}{\rotatebox[origin=c]{90}{\textbf{Corruptions}}}                                  & Shot                                  & \cellcolor{mygrey} \textbf{0.575}          & \cellcolor{mygrey} \textbf{0.236}          & 0.377          & \cellcolor{mygrey} \textbf{1.365}          & \cellcolor{mygrey} \textbf{0.063 }         & \cellcolor{mygrey} \textbf{0.425 }         & \multicolumn{1}{c}{\cellcolor{mygrey} \textbf{0.210}}          & \cellcolor{mygrey} \textbf{0.690}          \\
                                                     & Gaussian                                                     & 0.656          & \multicolumn{1}{c}{0.129}          & 0.192          & 0.794          & 0.036          & 0.344          & \multicolumn{1}{c}{0.112}          & 0.371          \\
                                                     & Impulse                                                          & 0.647          & \multicolumn{1}{c}{0.141}          & 0.245          & 0.843          & 0.037          & 0.353          & \multicolumn{1}{c}{0.123}          & 0.409          \\
                                                     & Speckle                                                         & 0.671          & \multicolumn{1}{c}{0.109}          & 0.198          & 0.578          & 0.026          & 0.329          & \multicolumn{1}{c}{0.094}          & 0.297          \\
                                                     & Defocus                                                        & 0.650          & \multicolumn{1}{c}{0.136}          & 0.272          & 0.691          & 0.031          & 0.350          & \multicolumn{1}{c}{0.119}          & 0.375          \\
                                                     & Zoom                                                        & 0.586          & \multicolumn{1}{c}{0.221}          & \cellcolor{mygrey} \textbf{0.516}          & 1.035          & 0.045          & 0.414          & \multicolumn{1}{c}{0.196}          & 0.616          \\
                                                     & Snow                                                         & 0.636          & \multicolumn{1}{c}{0.157}          & 0.350          & 0.682          & 0.030          & 0.364          & \multicolumn{1}{c}{0.136}          & 0.414          \\
                                                     & Bright                                                         & 0.704 & \multicolumn{1}{c}{0.066} & 0.112 & 0.279 & 0.013 & 0.296 & \multicolumn{1}{c}{0.054} & 0.151 \\
                                                     & Contrast                                                       & 0.663          & \multicolumn{1}{c}{0.120}          & 0.142          & 0.885          & 0.039          & 0.337          & \multicolumn{1}{c}{0.103}          & 0.368          \\
                                                     & Saturate                                                       & 0.681          & \multicolumn{1}{c}{0.097}          & 0.159          & 0.434          & 0.022          & 0.319          & \multicolumn{1}{c}{0.082}          & 0.241          \\
                                                     & Elastic                                                        & 0.680          & \multicolumn{1}{c}{0.098}          & 0.149          & 0.626          & 0.026          & 0.320          & \multicolumn{1}{c}{0.083}          & 0.277          \\
                                                     & Splatter                                                       & 0.674          & \multicolumn{1}{c}{0.106}          & 0.143          & 0.528          & 0.025          & 0.326          & \multicolumn{1}{c}{0.091}          & 0.268          \\
                                                     & Pixelate                                                       & 0.696          & \multicolumn{1}{c}{0.077}          & 0.126          & 0.478          & 0.019          & 0.304          & \multicolumn{1}{c}{0.064}          & 0.209          \\
                                                     & JPEG                                                       & 0.695          & \multicolumn{1}{c}{0.078}          & 0.134          & 0.402          & 0.017          & 0.305          & \multicolumn{1}{c}{0.065}          & 0.197  \\  \bottomrule     
\end{tabular}%
}
\caption{Comparison of accuracy, non-normalized aggregated error metrics, and VRE values of the models and corruptions introduced in this work. For the models, \textcolor{green}{$\uparrow$} indicates higher is more robust, and \textcolor{red}{$\downarrow$} indicates lower is more robust. The interpretation is reversed for visual corruptions \ie \textcolor{green}{$\uparrow$} indicate \emph{weaker} corruption strength and vice versa. \colorbox{mygrey}{\textbf{Bold}} denotes the most robust model or strongest corruption effect based on that particular metric. The VRE values have been computed using the default weight value \ie the weights are uniformly distributed among the normalized sub-metrics.}
\label{tab:metric}
\end{table}
    
    \subsection{Performance \vs Robustness}
    Can better performance be associated with greater robustness? \cref{tab:accVRETable} and \cref{fig:errorBar} highlight LLaVA-13B as the best-performing model, with the highest base accuracy and the least average error for most visual corruptions. However, \cref{tab:accVRETable} also shows ViLT having the least relative accuracy drop, and \cref{tab:metric,fig:metricComposition} consider ViLT as the most robust model based on VRE. Hence, better performance, in terms of accuracy, does not necessarily imply higher robustness. 

    \begin{figure*}[ht]
        \centering
        \begin{minipage}{0.85\textwidth}
        \centering
        \begin{subfigure}{0.157\textwidth}
            \centering
            \includegraphics[width=\textwidth, frame=0.2mm]{images/xai/i0s.jpg}
            \caption{}
        \end{subfigure}\hfill
        \begin{subfigure}{0.157\textwidth}
            \centering
            \includegraphics[width=\textwidth, frame=0.2mm]{images/xai/i1s.jpg}
            \caption{}
        \end{subfigure}\hfill
        \begin{subfigure}{0.157\textwidth}
            \centering
            \includegraphics[width=\textwidth, frame=0.2mm]{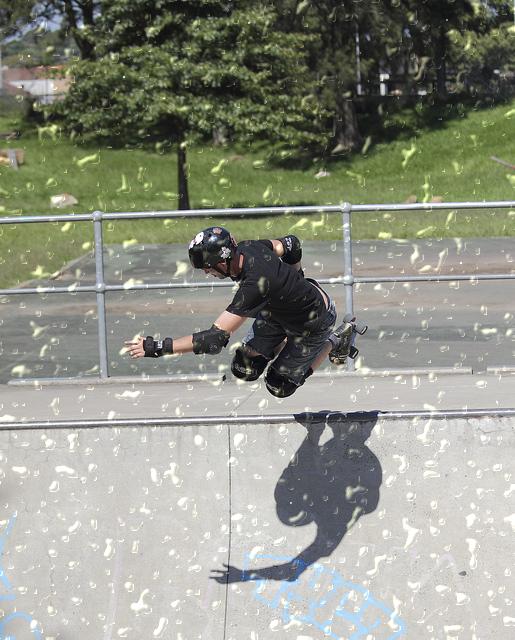}
            \caption{}
        \end{subfigure}\hfill
        \begin{subfigure}{0.157\textwidth}
            \centering
            \includegraphics[width=\textwidth, frame=0.2mm]{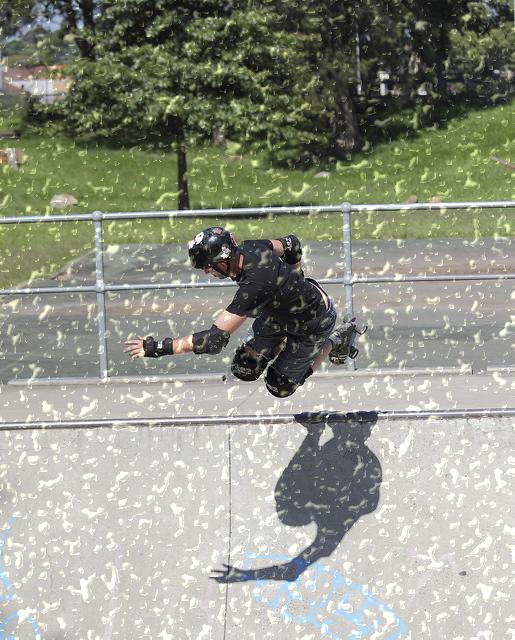}
            \caption{}
        \end{subfigure}\hfill
        \begin{subfigure}{0.157\textwidth}
            \centering
            \includegraphics[width=\textwidth, frame=0.2mm]{images/xai/i4s.jpg}
            \caption{}
        \end{subfigure}\hfill
        \begin{subfigure}{0.157\textwidth}
            \centering
            \includegraphics[width=\textwidth, frame=0.2mm]{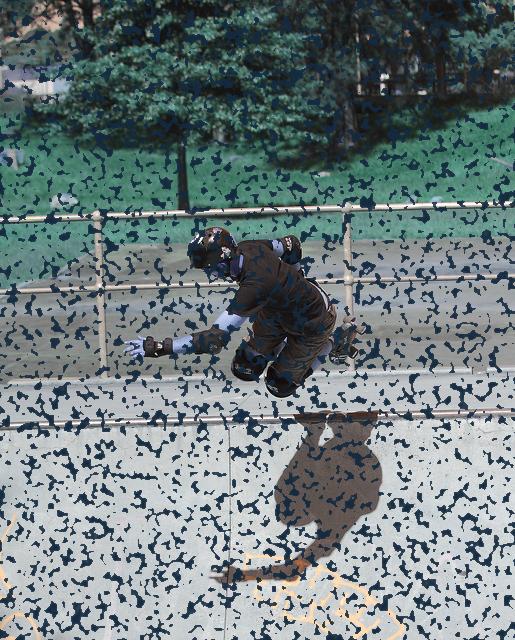}
            \caption{}
        \end{subfigure}
    
        \vspace{0.1cm}
        
        \begin{subfigure}{0.157\textwidth}
            \centering
            \includegraphics[width=\textwidth, frame=0.2mm]{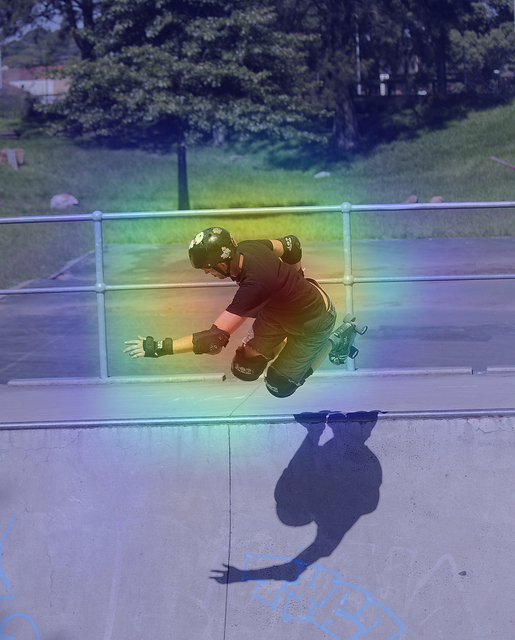}
            \caption{}
        \end{subfigure}\hfill
        \begin{subfigure}{0.157\textwidth}
            \centering
            \includegraphics[width=\textwidth, frame=0.2mm]{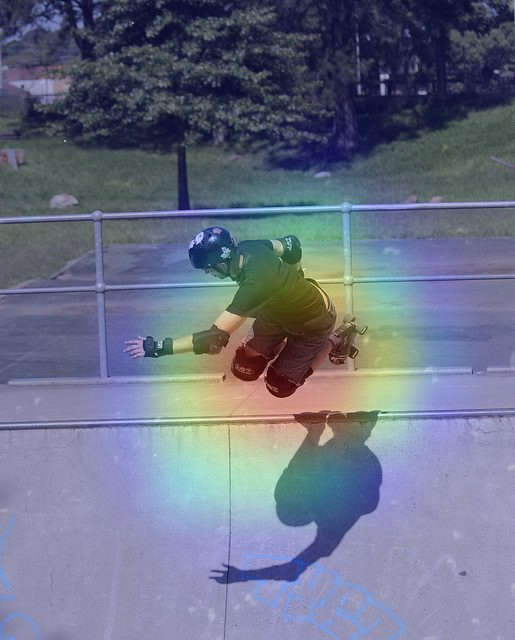}
            \caption{}
        \end{subfigure}\hfill
        \begin{subfigure}{0.157\textwidth}
            \centering
            \includegraphics[width=\textwidth, frame=0.2mm]{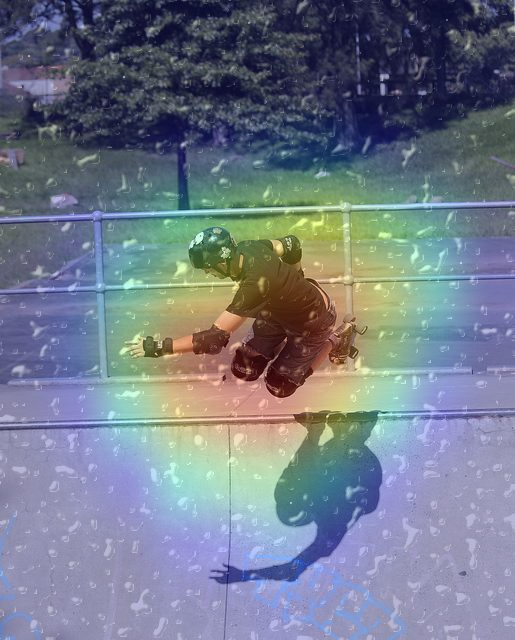}
            \caption{}
        \end{subfigure}\hfill
        \begin{subfigure}{0.157\textwidth}
            \centering
            \includegraphics[width=\textwidth, frame=0.2mm]{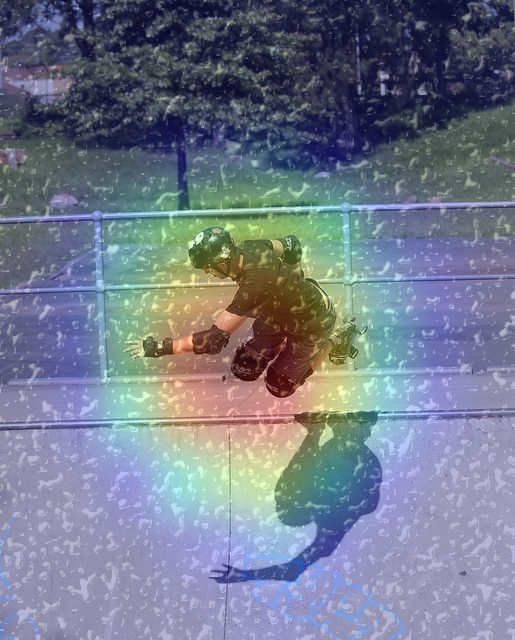}
            \caption{}
        \end{subfigure}\hfill
        \begin{subfigure}{0.157\textwidth}
            \centering
            \includegraphics[width=\textwidth, frame=0.2mm]{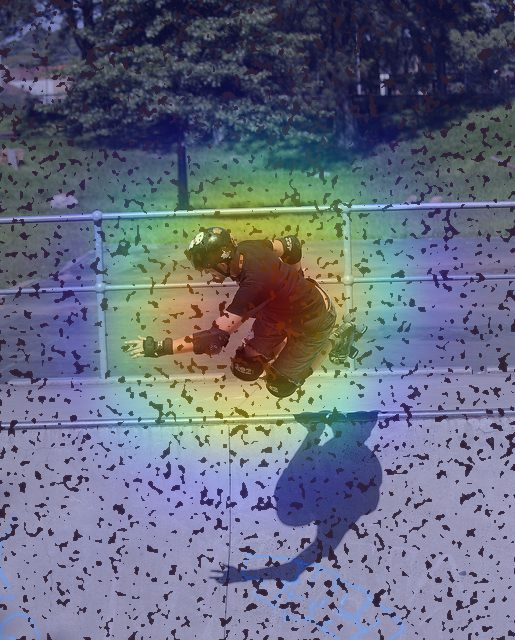}
            \caption{}
        \end{subfigure}\hfill
        \begin{subfigure}{0.157\textwidth}
            \centering
            \includegraphics[width=\textwidth, frame=0.2mm]{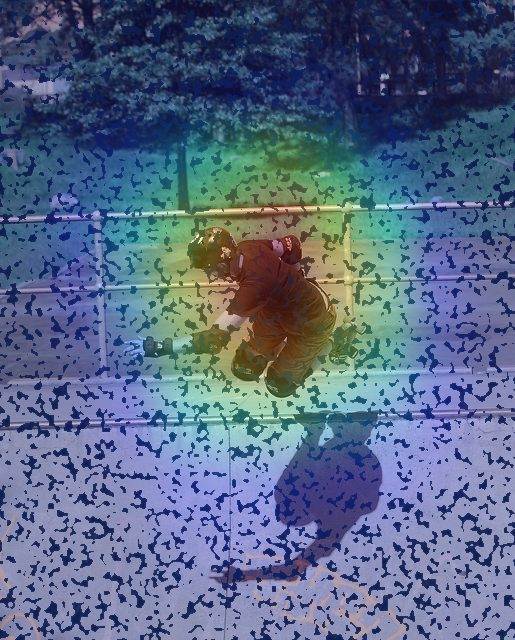}
            \caption{}
        \end{subfigure}
        
        \vspace{0.1cm}
    
        \begin{subfigure}{0.161\textwidth}
            \centering
            \includegraphics[width=\textwidth]{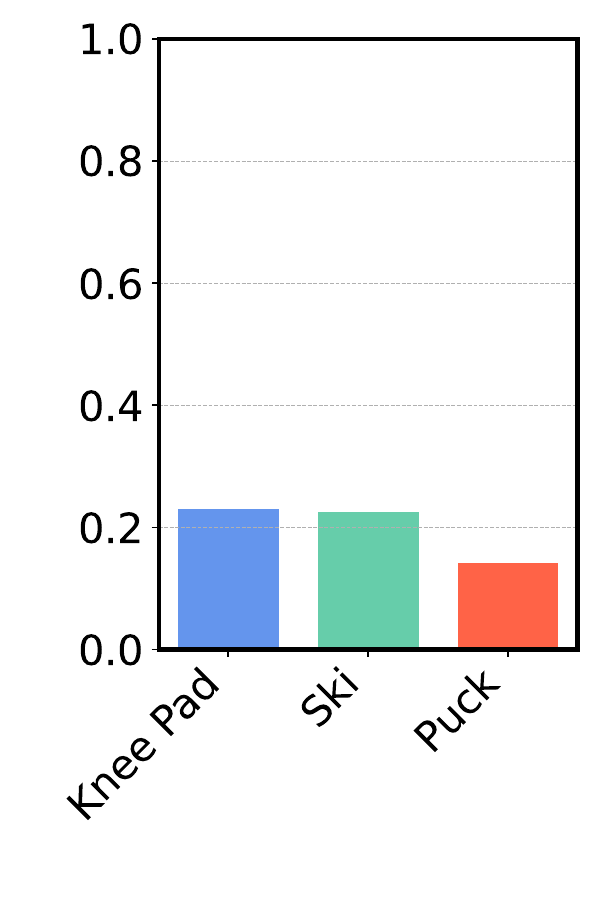}
            \caption{}
        \end{subfigure} \hfill
        \begin{subfigure}{0.161\textwidth}
            \centering
            \includegraphics[width=\textwidth]{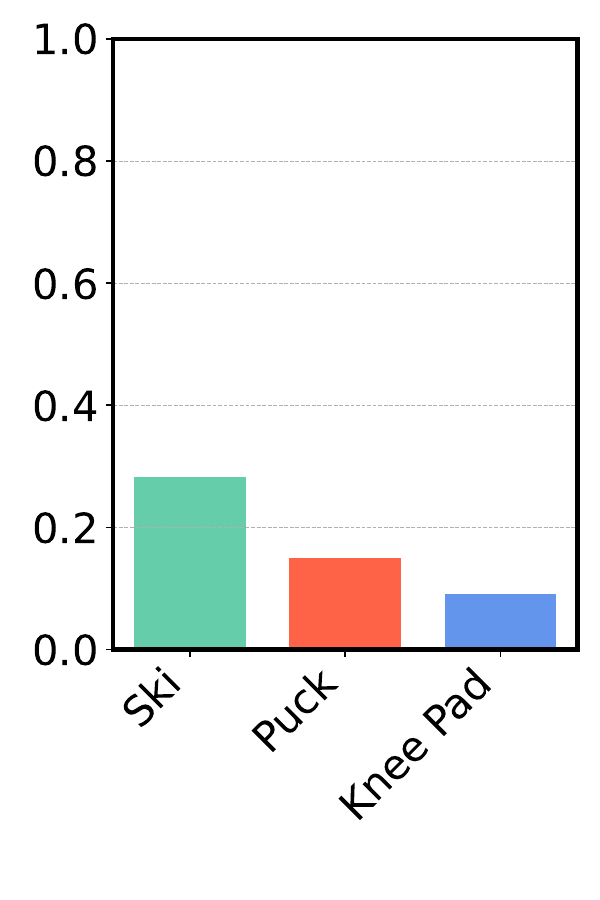}
            \caption{}
        \end{subfigure} \hfill
        \begin{subfigure}{0.161\textwidth}
            \centering
            \includegraphics[width=\textwidth]{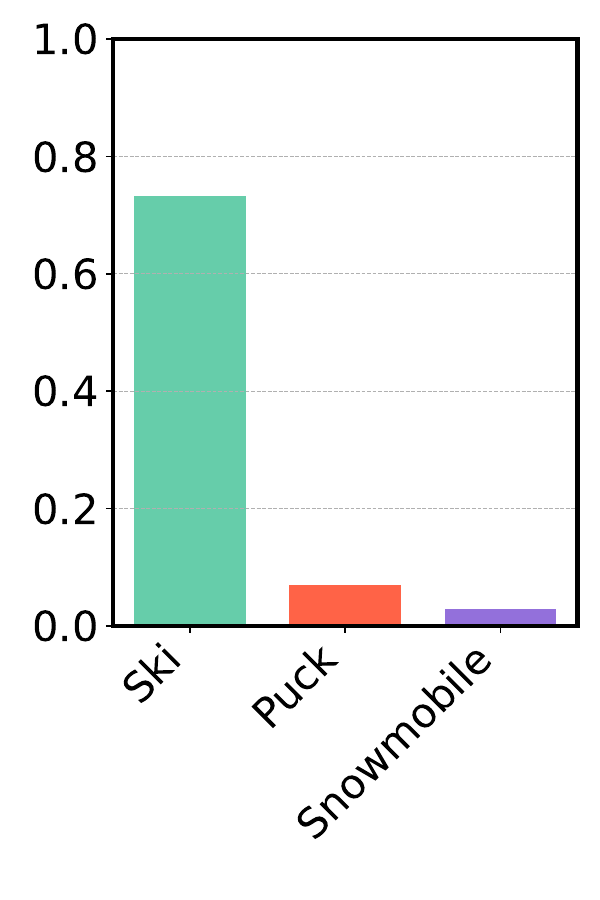}
            \caption{}
        \end{subfigure} \hfill
        \begin{subfigure}{0.161\textwidth}
            \centering
            \includegraphics[width=\textwidth]{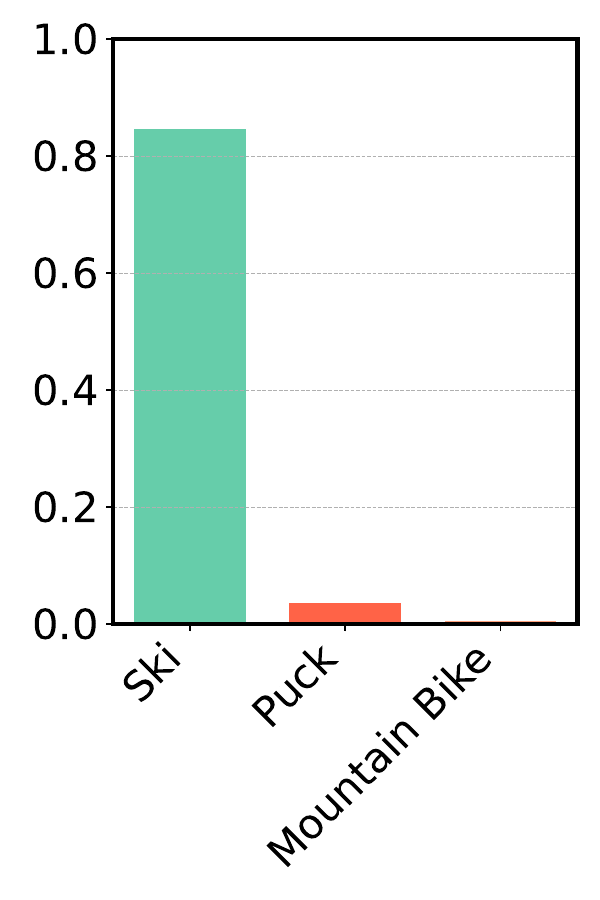}
            \caption{}
        \end{subfigure} \hfill
        \begin{subfigure}{0.161\textwidth}
            \centering
            \includegraphics[width=\textwidth]{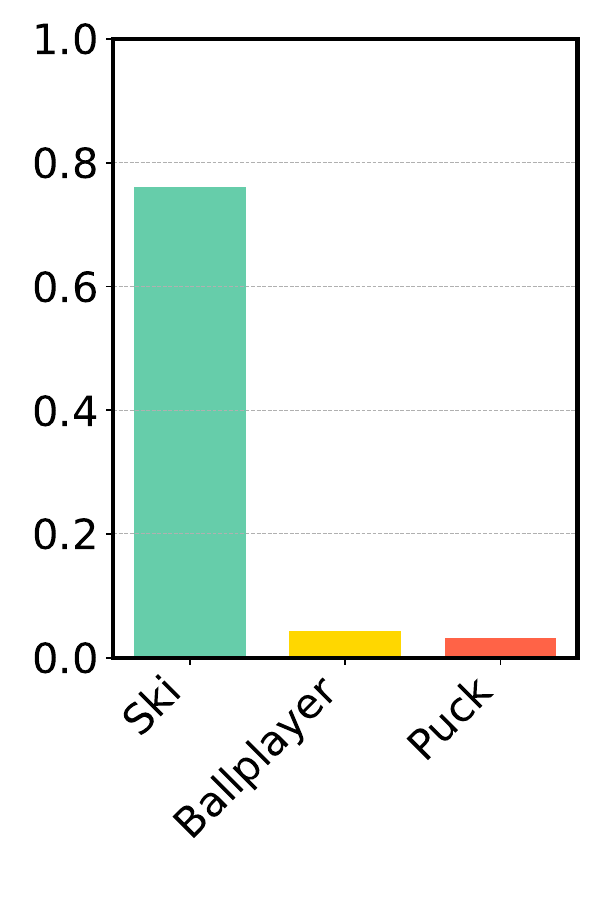}
            \caption{}
        \end{subfigure} \hfill
        \begin{subfigure}{0.161\textwidth}
            \centering
            \includegraphics[width=\textwidth]{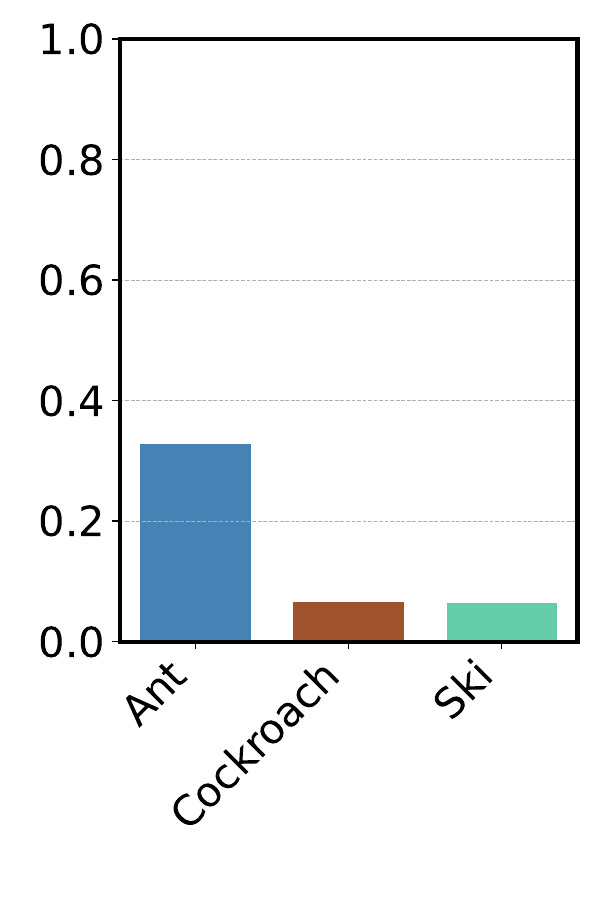}
            \caption{}
        \end{subfigure} 
        
        \end{minipage}
      \caption{(a)-(f) Set of images with increasing severity level when exposed to the splatter effect, (g)-(l) The corresponding Grad-CAM explanations \cite{selvaraju2017grad} employed by PNP \cite{tiong2022plug} showing variations in region localization, (m)-(r) The top 3 predictions and their associated probabilities, highlighting the inconsistencies in the corresponding caption predictions due to the visual corruption.}
      \label{fig:xai}
    \end{figure*}

 
\subsection{Strength of Visual Corruptions}
    \cref{tab:accVRETable,tab:metric} show an $6-24\%$ drop in accuracy due to visual corruptions. Based on the VRE value, we conclude that shot noise and zoom blur are the strongest visual corruptions. Other corruptions produce significantly lower VRE values, with brightness having the lowest value. Among the corruption classes, noise and blurring effects appear notably stronger, as illustrated in \cref{fig:metricComposition}.
    
 \subsection{Model Size and Robustness} 

    \cref{tab:modelParams} reveals that increasing model size does not necessarily translate into enhanced robustness. ViLT, the smallest model in terms of parameter count, is the most robust model based on VRE. However, excluding the zero-shot model PNP, we observe a marginal improvement in accuracy with the increased model size, suggesting robustness does not scale similarly to accuracy with model size. These findings highlight the importance of other factors in improving robustness, such as the training process, diversity and quality of the dataset, and architectural design.
    
   \definecolor{mygrey}{gray}{0.93}
\begin{table}[ht]
    \centering
    \begin{tabular}{cccc}
            \toprule
            \textbf{Model} & \textbf{\#Params}  & \textbf{$A_{v,0}$} \textcolor{green}{$\uparrow$} & \textbf{$\text{VRE}$} \textcolor{red}{$\downarrow$}\\ \midrule
            ViLT & 87.4M  & 0.713 & \cellcolor{mygrey}\textbf{0.221}              \\
            BLIP & 385M   & 0.782&   0.276            \\
            VLE & 378M    &0.771&     0.308         \\
            PnP & 223M   &0.648&   0.712            \\ 
            LLaVa-7B & 7B & 0.783 & 0.258\\
            LLaVa-13B & 13B & \cellcolor{mygrey}\textbf{0.798} & 0.290\\
            \bottomrule
        \end{tabular}
        \caption{The number of parameters, base accuracy, and VRE score of the six VQA models.}
        \label{tab:modelParams}
\end{table}
   
    \subsection{Model Explainability}
    The changes in localized regions in \cref{fig:xai} indicate that the corruption effects can affect the attention maps of the models and thereby alter the answer predictions. The Grad-CAM explanations \cite{selvaraju2017grad} are observed to be slightly shifted or dispersed when subjected to corruption effects, while the captioning predictions show severe fluctuations. Initially, the captioning module in PNP was confused between three likely predictions. At higher severity levels, it showed high confidence in a plausible class but ultimately ended with a confident yet absurdly wrong prediction. 

\begin{figure}[hb]
  \centering
  \includegraphics[width=0.45\textwidth]{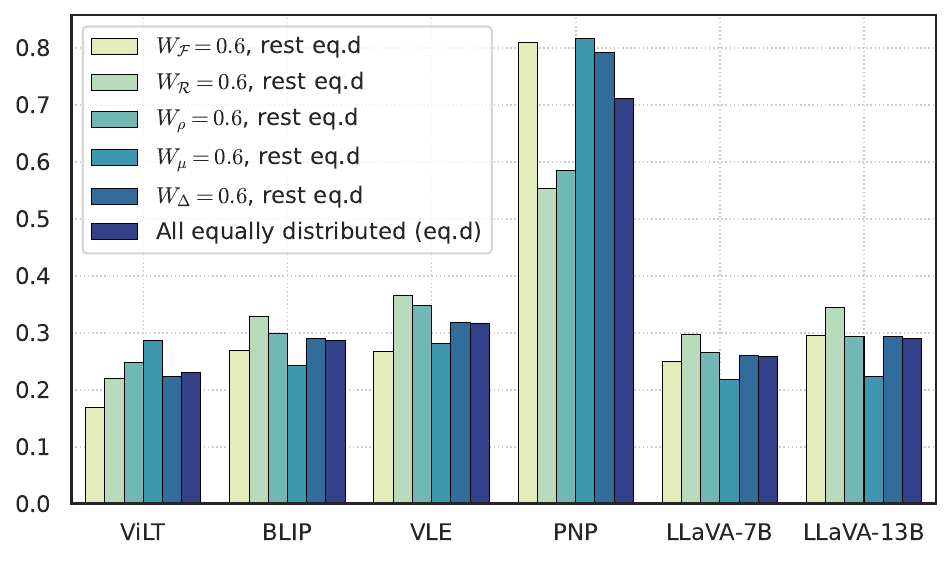}
  \captionsetup{skip=0pt}
  \caption{The change of $VRE_v$ with varying weight vectors. The weight of each sub-metric is taken as 0.6 while the other weights are equally distributed \ie 0.1 each. Lastly, all the weights are equally distributed among the sub-metrics \ie 0.2 each.}
  \label{fig:vreW}
\end{figure}

\subsection{Effect of Metric Weight Vector \texorpdfstring{$W$}{W} in VRE}
\label{sec:metricWeightEffectVRE}

The flexibility in setting the weight vector $W$ or preference scores $p$ enables us to focus on particular aspects of robustness, as illustrated in \cref{fig:vreW}. When the average error $\mu$ is prioritized using the higher weight of $0.6$, LLaVA-13B emerges as the most robust model, closely followed by the other standard VQA models. However, when any other sub-metric has a higher weight of $0.6$ or the weights are equally distributed, ViLT outperforms the other models. PNP consistently ranks last across all weight vectors. 

In practical scenarios, the evaluation metrics collectively provide an estimate of the robustness based on the use-case, as observed in Appendix \ref{app:robustnessScenarios}.


\section{Conclusion and Future Work}
 
We establish the first large-scale visual robustness benchmark for VQA by evaluating model robustness, assessing corruption strength, and quantifying these aspects using novel robustness evaluation metrics. While the proposed metrics are intuitively sound, an analytical justification is challenging due to the subjectivity inherent in robustness evaluation. Understanding the rationale behind prediction variations when subjected to visual corruption can be an interesting object of study in its own right.


We wish to explore several key directions, including the incorporation of textual corruption effects, consistency evaluation, and image denoising strategies. Although a few contemporary MLLMs have been included in this work, other larger models have demonstrated good zero-shot VQA performance \cite{achiam2023gpt} and can be included in future robustness analysis. Our work aims to lay the foundation for a universal Vision-Language evaluation suite that encompasses both visual and textual corruptions to develop robust VQA models. 


\section*{Code Availability}

Our code repository is available at: \url{https://github.com/ishmamt/Visual-Robustness}.

{\small
\bibliographystyle{ieee_fullname}
\bibliography{egbib}
}
\newpage
\clearpage
\setcounter{page}{1}
\maketitlesupplementary 
\appendix
\renewcommand{\thefigure}{A.\arabic{figure}}
\setcounter{figure}{0}
\renewcommand{\thetable}{A.\arabic{table}}
\setcounter{table}{0}

\section{Visual Corruption Function Details}
\label{app:visualCorruptionFunctions}
    \begin{figure*}[ht]
      \centering
        \includegraphics[width=\textwidth]{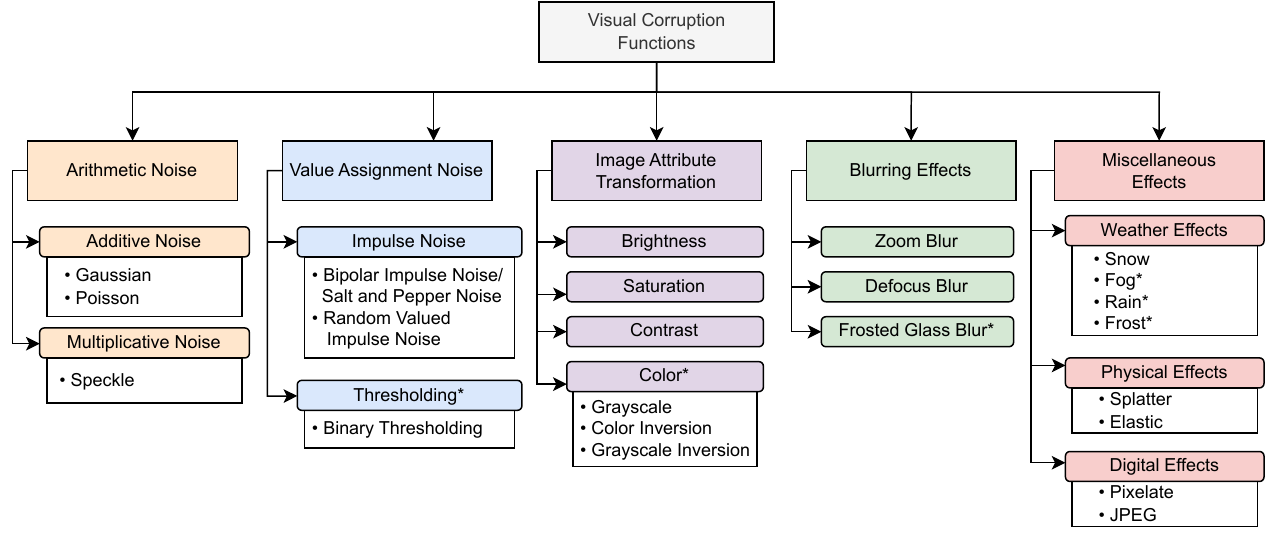}
    \caption{Comprehensive taxonomy of the visual corruption functions introduced in our work. The functions can be broadly categorized into five main classes similar to \cite{imgaug} which are further divided into multiple sub-classes, providing a detailed overview of the various types of realistic corruptions that can affect the quality of the image. * indicates that the corruption effects are included in our framework results of the corruption effects were not included in our work.}
    \label{fig:taxonomy}
    \end{figure*}

    \subsection{Arithmetic Noise}
    Arithmetic noise modifies the image by performing arithmetic operations \eg addition, multiplication, and negation on all of the color channels. A subcategory of arithmetic noise is \textbf{Additive Noise} which adds a particular value coming from a distribution $\mathcal{D}$ to every pixel in the image. Additive noise is implemented in the form of \textbf{Gaussian Noise} and \textbf{Poisson Noise}. Gaussian noise appears under low light conditions \cite{hendrycks2019benchmarking} and is one of the most common types of noise in telecommunications and digital image \cite{awad2019denoising}. Poisson noise or shot noise occurs due to the nature of light behaving as a quantized particle \cite{hasinoff2014photon}. 
    
    To define additive noise, we first define the random variable $X$ as $X_n \sim N(\mu, \sigma^2)$ where the probability distribution $N$ is defined as $f(x) = \frac{1}{\sigma\sqrt{2\pi}} e^{-\frac{(x-\mu)^2}{2\sigma^2}}$ and $X_p \sim P(\lambda)$ where the probability distribution $P$ is defined as $f(x) = \frac{\lambda^x}{x!} e^{-\lambda}$. The transformation function for additive noise can be generalized as $\mathcal{T}(r) = r + Y$ where $Y=X_n$ or $Y=X_p$ for Gaussian and Poisson noise respectively. The severity levels are defined by changing the parameter values of the aforementioned probability distributions.
    
    Another subcategory of arithmetic noise is \textbf{Multiplicative Noise} implemented in the form of \textbf{Speckle Noise} which can also be generalized as $\mathcal{T}(r) = r + r\cdot X_n$ where $X_n$ is the random variable $X_n \sim N(\mu, \sigma^2)$. Speckle noise is a common occurrence in medical and radar images \cite{maity2015comparative}. \textbf{Color Inversion}, a common digital image processing operation, performs subtraction \ie $\mathcal{T}(r) = r_{max} - r$. The deterministic function has a single severity level and can also be considered as an image attribute transformation function (\cref{sec:imageAttTrans}). The reader should note that the value of $\mathcal{T}(r)$ might fall outside the range $[r_{min},r_{max}]$ and requires clamping.
    
    \subsection{Value Assignment Noise}
    As the name suggests, the value assignment noise has a probability $p$ of assigning a particular value to a pixel, \ie $\mathcal{T}(r) = k$, on all the color channels. This noise is primarily implemented in the form of \textbf{Impulse Noise} which is typically one of the two types - bipolar impulse noise, commonly known as \textbf{Salt and Pepper Noise}, and \textbf{Random Valued Impulse Noise}. Salt and pepper noise takes one of two values, typically between the maximum intensity value $r_{max}$ and the minimum intensity value $r_{min}$ -- each with an equal probability $p$ of occurrence. Random-valued impulse noise takes a particular value from a range of values, typically $[r_{min},r_{max}]$, and follows a uniform distribution for the probabilistic occurrence of the values. A defective camera sensor might cause impulse noise during capturing and transmitting the image \cite{awad2019denoising, maity2018impulsive}. Another form of value assignment can take place in the form of \textbf{Thresholding} \ie the pixel will be assigned a binary value based on exceeding or subceeding a particular threshold value, $r_{thresh}$. \textbf{Binary Thresholding} is defined as $\mathcal{T}(r) = r_{max}$ if $r>r_{thresh}$, otherwise, $\mathcal{T}(r) = r_{min}$.

    \subsection{Image Attribute Transformation}
    \label{sec:imageAttTrans}
     Image attributes \eg, brightness, saturation, contrast, color properties, etc, are often modified to enhance the visual quality of the image \cite{gonzalez2009digital}. To modify the \textbf{Brightness}, we transform the image from the RGB color model to the HSV color model and add a positive or negative constant to the \textit{value} channel of the HSV image to increase or decrease the brightness. The function can be defined as $\mathcal{T}(v) = v + c$ where $v$ represents the value of the \textbf{value} channel and $c$ represents the additive constant. In real-life scenarios, lighting effects, luminance adjustment in digital displays, photographic effects, and other factors can cause an image to appear brighter or darker. By simulating these effects using the brightness function, our framework can test the visual robustness of VQA models under varying lighting and display conditions.
    
    \textbf{Saturation} refers to the purity of the colors in an image and can be used to enhance the quality of the image \ie the image will look visually appealing to a human observer \cite{gonzalez2009digital}. However, oversaturation might make the image look artificial to an observer and undersaturation might produce washed-out effects that can adversely affect the image quality. Changing the saturation is common in digital image processing to make the image look aesthetically pleasing or to reveal seemingly unseen features \cite{chiang2014color}. Saturation is changed by transforming the image from RGB to HSV color model, followed by modifying the \textit{saturation} channel value by multiplying and adding constants \ie$\mathcal{T}(v) = v \cdot c_1 + c_2$ where $c_1$ and $c_2$ represents the multiplicative and additive constants respectively which are set based on the severity of the noise.

    \textbf{Contrast} refers to the difference in color intensity values between different parts of the image \ie how well the details of an image are distinguishable \cite{gonzalez2009digital}. An image having a good level of contrast is more appealing to a viewer as it sets clear boundaries between various color intensities. On the contrary, low contrast creates difficulty in differentiating the details and hence, producing washed-out effects. Contrast enhancement is a common image-processing technique applied to spatial, frequency, and wavelet domains using contrast stretching, histogram equalization, etc. The contrast transformation is defined as,
    $\mathcal{T}(r) = (r - \mu)\cdot c + \mu$ where $\mu$ represents the average pixel intensity and $c$ represents the multiplicative constant. Similarly to arithmetic noise, the outputs of all image transformation functions are clamped.

    In real-world applications, grayscale images are prevalent due to constraints on representing the color information of a digital image. Several systems such as medical imaging, document scanning, and security work with grayscale images. On the other hand, systems like night vision, medical imaging, astronomy, etc. use color-inverted images. \textbf{Grayscale} can be categorized as a transformation function that modifies the color property of the image. Grayscale simply averages the intensity values over the color channels \ie $\mathcal{T}(r) = \mu^C$ where $\mu^C$ represents the average pixel intensity over the \textit{color} channel. \textbf{Color Inversion}, previously described as arithmetic noise, can be classified as an image attribute transformation function since it modifies the color property of an image. \textbf{Grayscale Inversion} is simply the combination of grayscale and color inversion; defined as $\mathcal{T}(r) = r_{max}-\mu_C$.    
    
    \subsection{Blurring Effects}
    Blurring effects are produced by convolving with an averaging filter and can be mathematically described as $\mathcal{T}(\mathcal{I}) = \mathcal{I} * K$ where $\mathcal{I}$ represents the digital image and $K$ represents the kernel and convolution operation for a 2D image is defined as 
    \begin{equation*}
        (\mathcal{I} * K)(x, y) = \sum_{i=0}^{M-1} \sum_{j=0}^{N-1} \mathcal{I}(i, j) \cdot K(x-i, y-j)
    \end{equation*}

    While Gaussian blur and median blur are the most common blurring functions, we shall define a few other blurring functions that have common real-life applications. \textbf{Defocus Blur} performs channel-wise convolution, and the function is defined as $\mathcal{T}(\mathcal{I}) = \mathcal{I} * K_{r}$ where $K_{r}$ is a disk kernel with radius $r$ that varies across severity levels. Defocus blur replicates the blurring effect in cameras when the subject is out of focus. \textbf{Zoom Blur} occurs due to rapid camera motion towards an object and \textbf{Frosted Glass Blur} imitates the appearance of an object while looking through frosted glass. Most of these effects do not have strict definitions and follows the implementation by \cite{hendrycks2019benchmarking, imgaug}.

    \subsection{Miscellaneous Effects}
    Apart from the previous transformation functions, weather effects can impose a particular weather condition on an image. At the time of writing this paper, our framework includes the \textbf{Snow Effect} only but we wish to include other effects like fog, frost, rain, and clouds in the future. We produce the snow effect by creating a snow layer following the normal distribution, then applying the zoom operation, followed by thresholding and motion blur. We use a blending function on the input image and a scaled grayscale version of this image, then add the snow layer and the rotated snow layer to the image to generate the final output of the snow effect.

    Some transformation functions try to create \textbf{Physical Effects} on the images. The \textbf{Splatter Effect} makes the image look like it has been splattered by paint or any form of liquid. The \textbf{Elastic Effect} simulates the effect of stretching or wrapping the image. Finally, we included a couple of transformation functions that replicate digitization effects. Digital images are discrete approximations of analog signals, thus various artifacts may remain from the conversion process. The \textbf{Pixelate Effect} is a visual effect that creates a mosaic-like appearance, similar to visible image pixels appearing due to lower resolutions, by downsampling and upsampling the image using bilinear interpolation. Pixelation is commonly used for stylistic purposes and censorship. \textbf{JPEG Compression Effect} tries to emulate the loss of image information due to JPEG compression \cite{santa2002jpeg}.

\section{Additional Evaluation Metric Details}
\label{app:evalMetrics}

\makenomenclature
\renewcommand\nomgroup[1]{%
  \item[\bfseries
  \ifstrequal{#1}{A}{General}{%
  \ifstrequal{#1}{B}{Accuracy Metrics}{%
  \ifstrequal{#1}{C}{Robustness Metrics}{%
  \ifstrequal{#1}{D}{Visual Robustness Error}{}}}}%
]}
\setlength{\nomitemsep}{-\parskip}

\nomenclature[A, 01]{\( v, \mathcal{V}, V \)}{Model, Set of Models, Number of Models}
\nomenclature[A, 02]{\( c, \mathcal{C}, C \)}{Corruption, Set of Corruptions, Number of Corruptions}
\nomenclature[A, 03]{\( l, \mathcal{L}, L \)}{Severity Level, Set of Severity Levels \{1,2,3,4,5\}, Number of Severity Levels}

\nomenclature[B, 01]{\( A_{v, c, l}\)}{Accuracy for model $v$, corruption $c$, and severity level $l$}
\nomenclature[B, 02]{\( A_{v, 0}  \)}{Base Accuracy for model $v$. Can be rewritten as $A_{v,c, 0}$}
\nomenclature[B, 03]{\( A_{v, c}  \)}{Average Accuracy for model $v$ and corruption $c$}
\nomenclature[B, 05]{\( A_v \)}{Corruption-average Accuracy for model $v$}
\nomenclature[B, 06]{\( A_c \)}{Model-average Accuracy for corruption $c$}
\nomenclature[B, 07]{\(A^{rel}_v\)}{Relative Accuracy Drop for model $v$}
\nomenclature[B, 08]{\(A^{rel}_c\)}{Relative Accuracy Drop for corruption $c$}

\nomenclature[C, 1]{\( E_{v, c, l} \)}{Error for model $v$, corruption $c$, and severity level $l$ }
\nomenclature[C, 2]{\( E_{v, 0} \)}{Base Error for model $v$,. Can be rewritten as $E_{v,c,0}$}
\nomenclature[C, 4]{\( \mathcal{M}_{v,c} \)}{Generalized Severity Aggregation Error Metric for model $v$ and corruption $c$}
\nomenclature[C, 5]{\( \mathcal{M'}_{v,c} \)}{Non-scaled Severity Aggregation Error Metric for model $v$ and corruption $c$}
\nomenclature[C, 6]{$\mathbb{M}$}{Set of Severity Aggregation Error Metrics $\{\mathcal{F},\mathcal{R},\rho,\mu,\delta\}$}
\nomenclature[C, 7]{\( \mathcal{M}_v \)}{Corruption Aggregation Error Metric for model $v$ }
\nomenclature[C, 8]{\( \mathcal{M}_c \)}{Model Aggregation Error Metric for corruption $c$ }

\nomenclature[C, 99999]{\( \mathcal{F} \)}{First-Drop}
\nomenclature[C, 9999]{\( \mathcal{R} \)}{Range of Error}
\nomenclature[C, 999]{\( \rho \)}{Error Rate}
\nomenclature[C, 99]{\( \mu \)}{Average Error}
\nomenclature[C, 9]{\( \delta \)}{Average Difference of Corruption Error}

\nomenclature[D, 01]{\(W_\mathcal{M} \)}{Weight assigned to Metric $\mathcal{M}$}
\nomenclature[D, 03]{\(p_\mathcal{M} \)}{Preference Score assigned to Metric $\mathcal{M}$}
\nomenclature[D, 07]{\(VRE_v   \)}{Visual Robustness Error for model $v$}
\nomenclature[D, 08]{\(VRE_c   \)}{Visual Robustness Error for corruption $c$}

\printnomenclature

\definecolor{mygrey}{gray}{0.93}
\definecolor{myred}{rgb}{1.0, 0.9, 0.9}
\definecolor{mygreen}{rgb}{0.9, 1.0, 0.9}
\begin{table*}[ht]

\centering

\resizebox{\textwidth}{!}{%
\begin{tabular}{cccccccccccccccc}
\toprule
\multirow{2}{*}{\textbf{Model}} & \multirow{2}{*}{\textbf{Lvl}} & \multicolumn{4}{c}{\textbf{Noise}}                               & \multicolumn{2}{c}{\textbf{Blur}}                & \multicolumn{1}{c}{\textbf{Weather}} & \multicolumn{3}{c}{\textbf{Image Attrribute}}               & \multicolumn{2}{c}{\textbf{Physical}}      & \multicolumn{2}{c}{\textbf{Digital}}     \\ \cmidrule(lr){3-6} \cmidrule(lr){7-8} \cmidrule(lr){9-9} \cmidrule(lr){10-12} \cmidrule(lr){13-14} \cmidrule(lr){15-16}
                       &                      & Shot        & Gaus  & Imp   & \multicolumn{1}{c}{Spec}  & Defo  & \multicolumn{1}{c}{Zoom}        & \multicolumn{1}{c}{Snow}    & Brig           & Cont  & \multicolumn{1}{c}{Sat}   & Elas  & \multicolumn{1}{c}{Spl}   & Pix            & JPEG           \\ \midrule
\multirow{6}{*}{ViLT}  & 0                    &  \multicolumn{14}{c}{\cellcolor{mygrey} \textbf{0.287}}                                                                                                                                                                                                                                      \\ 
                       & 1                    & 0.326       & 0.290 & 0.297 & \multicolumn{1}{c}{0.290} & 0.295 & \multicolumn{1}{c}{{\cellcolor{myred}  \textbf{0.392}}} & \multicolumn{1}{c}{0.341}   & 0.287          & 0.290 & \multicolumn{1}{c}{0.290} & 0.292 & \multicolumn{1}{c}{0.286} & \cellcolor{mygreen} \textbf{0.280} & 0.282          \\
                       & 2                    & 0.374       & 0.304 & 0.314 & \multicolumn{1}{c}{0.293} & 0.307 & \multicolumn{1}{c}{{\cellcolor{myred}  \textbf{0.439}}} & \multicolumn{1}{c}{0.395}   & 0.291          & 0.298 & \multicolumn{1}{c}{0.307} & 0.296 & \multicolumn{1}{c}{0.317} & \cellcolor{mygreen} \textbf{0.281} & 0.285          \\
                       & 3                    & 0.460       & 0.329 & 0.334 & \multicolumn{1}{c}{0.321} & 0.332 & \multicolumn{1}{c}{{\cellcolor{myred}  \textbf{0.472}}} & \multicolumn{1}{c}{0.424}   & 0.301          & 0.317 & \multicolumn{1}{c}{0.323} & 0.305 & \multicolumn{1}{c}{0.348} & \cellcolor{mygreen} \textbf{0.284} & 0.290          \\
                       & 4                    & {\cellcolor{myred}  \textbf{0.526}} & 0.369 & 0.380 & \multicolumn{1}{c}{0.336} & 0.359 & \multicolumn{1}{c}{0.503}       & \multicolumn{1}{c}{0.458}   & 0.312          & 0.407 & \multicolumn{1}{c}{0.356} & 0.329 & \multicolumn{1}{c}{0.370} & \cellcolor{mygreen} \textbf{0.291} & 0.303          \\
                       & 5                    & {\cellcolor{myred}  \textbf{0.569}} & 0.436 & 0.437 & \multicolumn{1}{c}{0.364} & 0.387 & \multicolumn{1}{c}{0.525}       & \multicolumn{1}{c}{0.475}   & 0.328          & 0.567 & \multicolumn{1}{c}{0.359} & 0.366 & \multicolumn{1}{c}{0.408} & 0.340          & \cellcolor{mygreen} \textbf{0.322} \\ \midrule
\multirow{6}{*}{BLIP}  & 0                    & \multicolumn{14}{c}{ \cellcolor{mygrey} \textbf{0.218}}                                                                                                                                                                                                                                      \\
                       & 1                    & 0.281       & 0.238 & 0.253 & \multicolumn{1}{c}{0.236} & 0.254 & \multicolumn{1}{c}{{\cellcolor{myred}  \textbf{0.337}}} & \multicolumn{1}{c}{0.286}   & \cellcolor{mygreen} \textbf{0.228} & 0.237 & \multicolumn{1}{c}{0.240} & 0.233 & \multicolumn{1}{c}{0.233} & 0.238          & 0.239          \\
                       & 2                    & 0.328       & 0.254 & 0.270 & \multicolumn{1}{c}{0.244} & 0.276 & \multicolumn{1}{c}{{\cellcolor{myred}  \textbf{0.377}}} & \multicolumn{1}{c}{0.322}   & \cellcolor{mygreen} \textbf{0.236} & 0.248 & \multicolumn{1}{c}{0.261} & 0.251 & \multicolumn{1}{c}{0.262} & 0.248          & 0.251          \\
                       & 3                    & 0.410       & 0.283 & 0.287 & \multicolumn{1}{c}{0.275} & 0.308 & \multicolumn{1}{c}{{\cellcolor{myred}  \textbf{0.411}}} & \multicolumn{1}{c}{0.325}   & \cellcolor{mygreen} \textbf{0.246} & 0.267 & \multicolumn{1}{c}{0.278} & 0.270 & \multicolumn{1}{c}{0.292} & 0.257          & 0.258          \\
                       & 4                    & {\cellcolor{myred}  \textbf{0.471}} & 0.324 & 0.332 & \multicolumn{1}{c}{0.299} & 0.338 & \multicolumn{1}{c}{0.441}       & \multicolumn{1}{c}{0.358}   & \cellcolor{mygreen} \textbf{0.260} & 0.320 & \multicolumn{1}{c}{0.305} & 0.280 & \multicolumn{1}{c}{0.299} & 0.276          & 0.285          \\
                       & 5                    & {\cellcolor{myred}  \textbf{0.533}} & 0.389 & 0.394 & \multicolumn{1}{c}{0.329} & 0.367 & \multicolumn{1}{c}{0.461}       & \multicolumn{1}{c}{0.373}   & \cellcolor{mygreen} \textbf{0.275} & 0.414 & \multicolumn{1}{c}{0.315} & 0.366 & \multicolumn{1}{c}{0.332} & 0.349          & 0.316          \\ \midrule
\multirow{6}{*}{VLE}   & 0                    & \multicolumn{14}{c}{\cellcolor{mygrey} \textbf{0.229}}                                                                                                                                                                                                                                      \\  
                       & 1                    & 0.311       & 0.250 & 0.261 & \multicolumn{1}{c}{0.245} & 0.271 & \multicolumn{1}{c}{{\cellcolor{myred}  \textbf{0.366}}} & \multicolumn{1}{c}{0.266}   & \cellcolor{mygreen} \textbf{0.230} & 0.235 & \multicolumn{1}{c}{0.238} & 0.242 & \multicolumn{1}{c}{0.233} & 0.236          & 0.248          \\
                       & 2                    & 0.387       & 0.271 & 0.290 & \multicolumn{1}{c}{0.260} & 0.296 & \multicolumn{1}{c}{{\cellcolor{myred}  \textbf{0.410}}} & \multicolumn{1}{c}{0.312}   & \cellcolor{mygreen} \textbf{0.233} & 0.241 & \multicolumn{1}{c}{0.242} & 0.249 & \multicolumn{1}{c}{0.255} & 0.238          & 0.267          \\
                       & 3                    & {\cellcolor{myred}  \textbf{0.499}} & 0.324 & 0.323 & \multicolumn{1}{c}{0.309} & 0.347 & \multicolumn{1}{c}{0.425}       & \multicolumn{1}{c}{0.314}   & \cellcolor{mygreen} \textbf{0.241} & 0.253 & \multicolumn{1}{c}{0.265} & 0.287 & \multicolumn{1}{c}{0.275} & 0.256          & 0.275          \\
                       & 4                    & {\cellcolor{myred}  \textbf{0.567}} & 0.400 & 0.404 & \multicolumn{1}{c}{0.356} & 0.392 & \multicolumn{1}{c}{0.468}       & \multicolumn{1}{c}{0.338}   & \cellcolor{mygreen} \textbf{0.249} & 0.307 & \multicolumn{1}{c}{0.286} & 0.291 & \multicolumn{1}{c}{0.278} & 0.280          & 0.316          \\
                       & 5                    & {\cellcolor{myred}  \textbf{0.606}} & 0.487 & 0.478 & \multicolumn{1}{c}{0.396} & 0.420 & \multicolumn{1}{c}{0.489}       & \multicolumn{1}{c}{0.350}   & \cellcolor{mygreen} \textbf{0.267} & 0.431 & \multicolumn{1}{c}{0.301} & 0.411 & \multicolumn{1}{c}{0.313} & 0.387          & 0.373          \\ \midrule
\multirow{6}{*}{PNP}   & 0                    & \multicolumn{14}{c}{\cellcolor{mygrey} \textbf{0.352}}                                                                                                                                                                                                                                      \\ 
                       & 1                    & 0.585       & 0.575 & 0.580 & \multicolumn{1}{c}{0.575} & 0.579 & \multicolumn{1}{c}{{\cellcolor{myred}  \textbf{0.603}}} & \multicolumn{1}{c}{0.590}   & \cellcolor{mygreen} \textbf{0.570} & 0.576 & \multicolumn{1}{c}{0.576} & 0.572 & \multicolumn{1}{c}{0.574} & 0.573          & 0.572          \\
                       & 2                    & 0.599       & 0.578 & 0.583 & \multicolumn{1}{c}{0.576} & 0.584 & \multicolumn{1}{c}{{\cellcolor{myred}  \textbf{0.612}}} & \multicolumn{1}{c}{0.603}   & 0.574          & 0.580 & \multicolumn{1}{c}{0.590} & 0.578 & \multicolumn{1}{c}{0.585} & \cellcolor{mygreen} \textbf{0.574} & 0.575          \\
                       & 3                    & {\cellcolor{myred}  \textbf{0.626}} & 0.585 & 0.586 & \multicolumn{1}{c}{0.586} & 0.591 & \multicolumn{1}{c}{0.621}       & \multicolumn{1}{c}{0.606}   & 0.578          & 0.586 & \multicolumn{1}{c}{0.593} & 0.582 & \multicolumn{1}{c}{0.590} & \cellcolor{mygreen} \textbf{0.575} & 0.577          \\
                       & 4                    & {\cellcolor{myred}  \textbf{0.650}} & 0.599 & 0.601 & \multicolumn{1}{c}{0.588} & 0.599 & \multicolumn{1}{c}{0.629}       & \multicolumn{1}{c}{0.615}   & 0.580          & 0.606 & \multicolumn{1}{c}{0.607} & 0.583 & \multicolumn{1}{c}{0.598} & \cellcolor{mygreen} \textbf{0.580} & 0.582          \\
                       & 5                    & {\cellcolor{myred}  \textbf{0.681}} & 0.616 & 0.615 & \multicolumn{1}{c}{0.601} & 0.611 & \multicolumn{1}{c}{0.642}       & \multicolumn{1}{c}{0.619}   & \cellcolor{mygreen} \textbf{0.588} & 0.638 & \multicolumn{1}{c}{0.610} & 0.608 & \multicolumn{1}{c}{0.608} & 0.599          & 0.590  \\ \midrule
       \multirow{6}{*}{LLaVA-7B} & 0 & \multicolumn{14}{c}{\cellcolor{mygrey}\textbf{0.217}}                                                                                                               \\
                          & 1 & 0.298          & 0.246 & 0.262 & 0.256 & 0.275 & \cellcolor{myred}\textbf{0.301} & 0.296 & 0.217
                          & 0.226 & 0.231 & 0.225 & 0.229 & \cellcolor{mygreen}\textbf{0.217}          & 0.218 \\
                          & 2 & 0.341          & 0.267 & 0.280  & 0.264 & 0.291 & \cellcolor{myred}\textbf{0.348} & 0.319 & 0.221          & 0.239 & 0.245 & 0.234 & 0.257 & \cellcolor{mygreen}\textbf{0.217} & 0.220  \\
                          & 3 & \cellcolor{myred}\textbf{0.396} & 0.281 & 0.303 & 0.273 & 0.313 & 0.372          & 0.322 & 0.230           & 0.249 & 0.271 & 0.249 & 0.291 & \cellcolor{mygreen}\textbf{0.222} & 0.225 \\
                          & 4 & \cellcolor{myred}\textbf{0.453} & 0.319 & 0.341 & 0.312 & 0.338 & 0.413          & 0.347 & 0.243          & 0.315 & 0.294 & 0.277 & 0.299 & \cellcolor{mygreen}\textbf{0.229} & 0.231 \\
                          & 5 & \cellcolor{myred}\textbf{0.542} & 0.374 & 0.407 & 0.346 & 0.369 & 0.452          & 0.361 & \cellcolor{mygreen}\textbf{0.259} & 0.397 & 0.301 & 0.349 & 0.327 & 0.283          & 0.267 \\ \midrule
                          \multirow{6}{*}{LLaVA-13B} & 0 & \multicolumn{14}{c}{\cellcolor{mygrey}\textbf{0.202}}                                                                                                               \\
                           & 1 & 0.291          & 0.241 & 0.258 & 0.245 & 0.271 & \cellcolor{myred}\textbf{0.300} & 0.283 & \cellcolor{mygreen}\textbf{0.203} & 0.212 & 0.223 & 0.220 & 0.220 & 0.208          & 0.204 \\
                           & 2 & 0.333          & 0.265 & 0.279 & 0.259 & 0.293 & \cellcolor{myred}\textbf{0.348} & 0.304 & \cellcolor{mygreen}\textbf{0.205}          & 0.225 & 0.238 & 0.223 & 0.255 & 0.209 & 0.209 \\
                           & 3 & \cellcolor{myred}\textbf{0.378} & 0.282 & 0.305 & 0.267 & 0.315 & 0.371          & 0.310 & \cellcolor{mygreen}\textbf{0.212}          & 0.234 & 0.264 & 0.239 & 0.287 & 0.219 & 0.215 \\
                           & 4 & \cellcolor{myred}\textbf{0.447} & 0.323 & 0.347 & 0.304 & 0.339 & 0.411          & 0.341 & 0.234          & 0.312 & 0.291 & 0.271 & 0.298 & \cellcolor{mygreen}\textbf{0.225} & 0.229 \\
                           & 5 & \cellcolor{myred}\textbf{0.541} & 0.376 & 0.408 & 0.336 & 0.372 & 0.449          & 0.358 & \cellcolor{mygreen}\textbf{0.249} & 0.386 & 0.297 & 0.338 & 0.325 & 0.272          & 0.258\\
                       \bottomrule       
\end{tabular}%
}

\caption{Error values of the model across various severity levels for different visual corruption effects. \colorbox{mygreen}{\textbf{Green}} and \colorbox{myred}{\textbf{red}} indicate the minimum and maximum error values of that particular corruption and severity level respectively. The minimum values are observed at only three corruption effects -- brightness, pixelate, and JPEG, while the maximum values are observed at zoom blur and shot noise only.} 
\label{tab:errorVRE}
\end{table*}

\subsection{First-Drop Details} The rationale behind using \emph{relative difference} instead of \emph{difference} is that the difference between level-1 and level-0 errors will depend on the model's base accuracy. The relative difference does not have such a dependency and can better capture the error when the model is introduced to corruption effects while encapsulating the variations and decoupling the base accuracy from the equation.

The word \textit{drop} signifies the drop or decrease in accuracy due to introducing the visual corruption. The term might be misleading since, when calculating robustness error, the addition of corruption increases the error. Higher first-drop scores for a model-corruption pair indicate that the model is more prone to lower levels of that particular corruption. 

\subsection{Range of Error Details} A clear distinction is made between the maximum error value and the error value at the highest severity level in \cref{eq:errorRange} as the error value at the highest severity level does not imply that it has the maximum error value.
    \begin{figure*}[ht!]
      \centering
      \includegraphics[width = 0.9\textwidth]{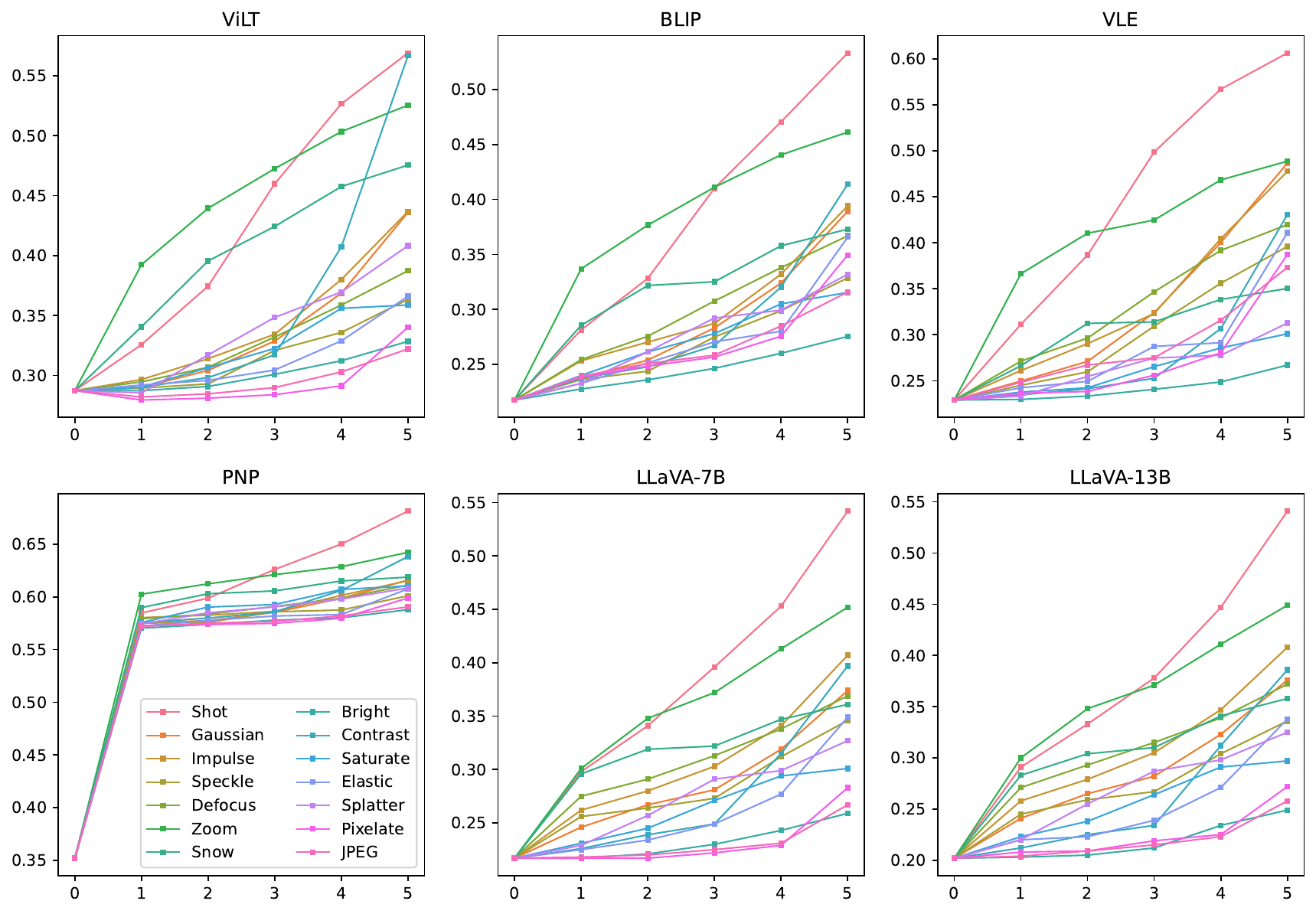}
      \caption{Error trends over severity levels. The standard VQA models -- ViLT \cite{kim2021vilt}, BLIP \cite{li2022blip}, VLE \cite{iflytekVLE}, and LLaVA \cite{liu2024improved} exhibit a somewhat linear rise while the zero-shot VQA model PNP \cite{tiong2022plug} shows an initial sharp rise followed by a linear trend. Logarithmic trend lines are observed by a few corruption functions \eg zoom blur.}
      \label{fig:modelArchi}
    \end{figure*}

\subsection{Robustness Evaluation Scenarios using VRE}
\label{app:robustnessScenarios}
\textbf{Scenario 1.}~~We assume the VQA model is deployed in an environment where minor levels of visual corruption occur \eg slight changes in weather or lighting conditions. This scenario is well-suited for the first-drop evaluation metric, which captures robustness at lower severity levels. The average error, which provides an overall estimation of performance, can also be used. VRE can be tuned by assigning more weight to the first drop and average error metrics.

\textbf{Scenerio 2.}~~We consider that the VQA model is experiencing corruption effects that gradually increase in intensity \eg the images are getting blurrier or brighter over time. In this case, the range of error and error rate can jointly evaluate the robustness. VRE can be similarly tuned to prioritize the aforementioned metrics.

\section{Robustness across Question Types}
\label{app:robustnessQuestionTypes}
\begin{figure*}[ht!]
  \centering
  \includegraphics[width = 0.9\textwidth]{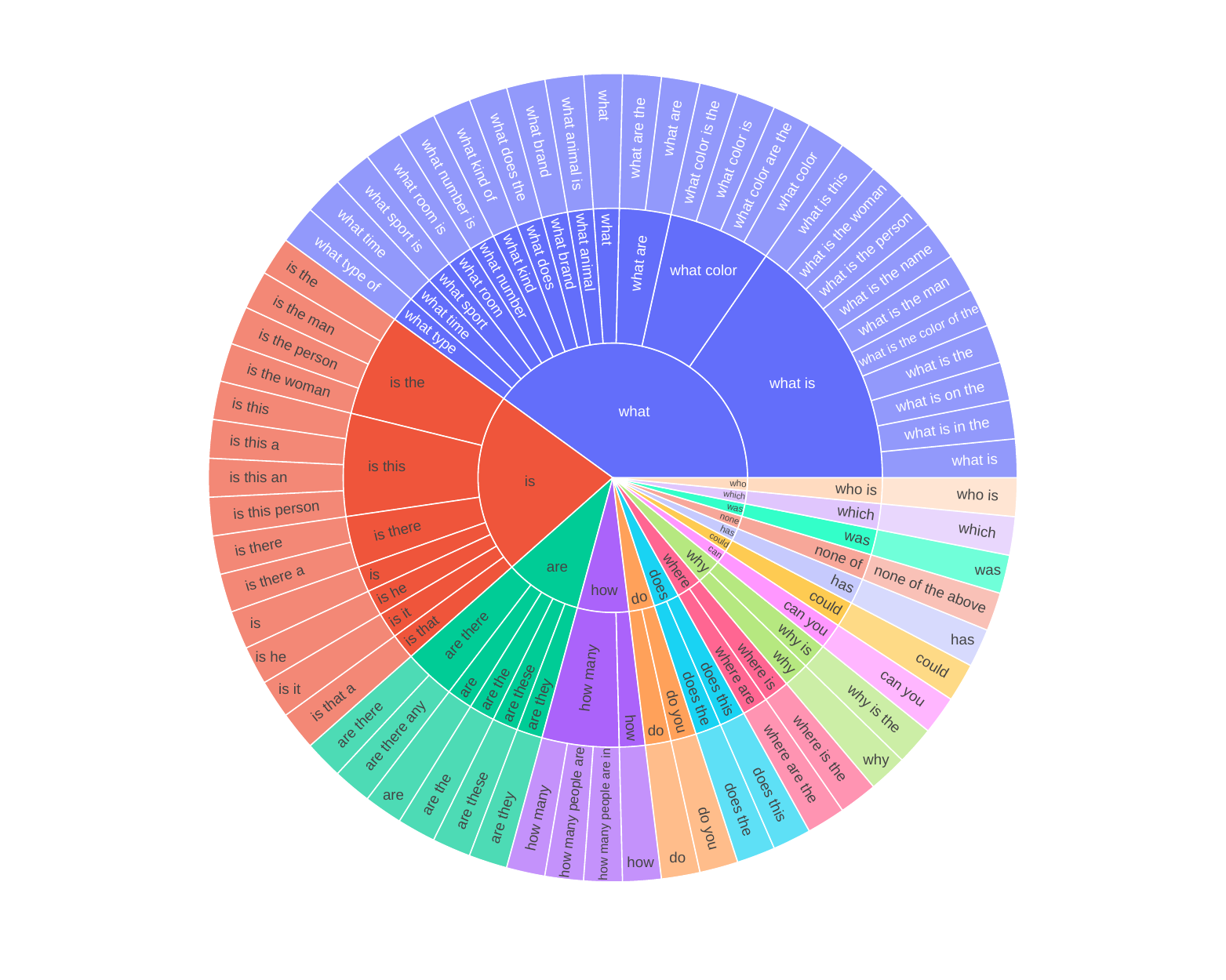}
  \caption{Breakdown of different question types from the VQAv2 \cite{goyal2017making} dataset. The most frequent questions start with ``What'' followed by ``Is'', and ``Are''.}
  \label{fig:questionType}
\end{figure*}

\begin{table*}[ht]
\centering
\begin{tabular}{lccccccc}
\toprule
\textbf{Corruption} & \textbf{Is this} & \textbf{What} & \textbf{What is} & \textbf{Is/are the} & \textbf{How many} & \textbf{What color is} & \textbf{What kind of} \\ \midrule
\textbf{Shot}       & 0.731    & 0.363 & 0.312    & 0.711       & 0.398     & 0.557          & 0.412         \\
\textbf{Gaussian}   & 0.812    & 0.454 & 0.432    & 0.764       & 0.466     & 0.698          & 0.528         \\
\textbf{Impulse}    & 0.817    & 0.445 & 0.428    & 0.762       & 0.453     & 0.691          & 0.525         \\
\textbf{Speckle}    & 0.828    & 0.476 & 0.457    & 0.771       & 0.483     & 0.722          & 0.546         \\
\textbf{Defocus}    & 0.812    & 0.444 & 0.412    & 0.766       & 0.442     & 0.731          & 0.524         \\
\textbf{Zoom}       & 0.731    & 0.369 & 0.297    & 0.683       & 0.350     & 0.663          & 0.413         \\
\textbf{Snow}       & 0.771    & 0.428 & 0.389    & 0.741       & 0.431     & 0.656          & 0.491         \\
\textbf{Brightness} & 0.843    & 0.494 & 0.489    & 0.798       & 0.561      & 0.718          & 0.579         \\
\textbf{Contrast}   & 0.814    & 0.453 & 0.433    & 0.769       & 0.457     & 0.659          & 0.518         \\
\textbf{Saturation} & 0.833    & 0.473 & 0.465    & 0.799       & 0.499     & 0.597          & 0.545         \\
\textbf{Elastic}    & 0.811    & 0.462 & 0.451    & 0.774       & 0.466     & 0.758          & 0.546         \\
\textbf{Pixelate}   & 0.849    & 0.480 & 0.467    & 0.783       & 0.486     & 0.756          & 0.557         \\
\textbf{JPEG}       & 0.848    & 0.481 & 0.473    & 0.780       & 0.480     & 0.737          & 0.553         \\
\textbf{Spatter}    & 0.815    & 0.460 & 0.422    & 0.772       & 0.479     & 0.720          & 0.531        \\ \bottomrule
\end{tabular}
\caption{Average accuracy across different types of questions for the visual corruption functions. The models primarily struggle with the ``How many" and ``What is" questions.}
\end{table*}

The question category composition of our subsample of the VQAv2 dataset\cite{goyal2017making} has been explored in fig-\ref{fig:questionType}, where the ``What'' type questions are most frequently observed. Following this, most of the questions encountered by the VQA models are asked to classify certain objects or their properties, for instance, ``What is'', ``What color'', and ``What animal''. The next two most encountered question types, ``Is'' and ``Are'', ask the model to verify the existence or absence of something in the image. These questions are inherently biased to allude to an object or an action being present in the image itself. Thus, we conclude that the questions from the VQAv2 dataset are not diverse and are somewhat lacking in judging the model's critical thinking capabilities. Questions starting with ``How to'', ``Why is'', ``Who'', and more are rarer while the dataset lacks questions challenging the counting ability of the model \eg questions with ``How many''. 

\section{Discussion}  
\label{app:discussion}
\subsection{The Necessity of Robustness}

A model selected based on high average accuracy or low average error may provide precise and correct predictions under ideal conditions but becomes susceptible to producing erroneous outputs when faced with variations, uncertainties, or adversarial inputs. However, a robust model selected based on multiple aspects exhibits a higher level of resilience and generalization, capable of performing consistently across a wide range of inputs, even in the face of perturbations or challenging scenarios. The increased robustness might come at the cost of sacrificing accuracy, as the model adopts a more conservative or cautious approach to minimize error.

The trade-off between accuracy and robustness is crucial to consider when developing machine learning models for various applications. Different contexts and use cases may require varying degrees of emphasis on accuracy and robustness. For instance, in safety-critical systems, such as autonomous vehicles or medical diagnosis, robustness takes precedence over accuracy to ensure reliable performance even in uncertain or unpredictable situations. However, in tasks where precision and correctness are paramount, sacrificing some robustness may be acceptable to achieve higher accuracy.
\begin{table*}[ht]
\centering

\label{tab:questionError}
    \begin{tabular}{llll}
    \toprule
    \textbf{Problem} & \textbf{Questions} & \textbf{Ground Truth} & \textbf{Predictions}\\
    \midrule
  \multirow{3}{*}{Miscolor}
  & What is the bike's color?& blue & black\\
  & What color is the sky? & blue & gray\\
  & What is the color of the soap? & yellow & white\\
  
  \midrule
  \multirow{3}{*}{Undercount} & How many spoons are there? & 2 & 1\\
  & How many people are there? & 5 & 3\\
  & How many kites are up? & 4 & 3\\
  
  \midrule
  \multirow{3}{*}{Misclassify} & What is she eating? & sandwich & cake\\
  & What is the weather like? & sunny & cloudy\\
  & What game is this? & baseball & soccer\\

  \midrule
  \multirow{3}{*}{Blindness} & What's on the television? & baby & nothing\\
  & Are the women selling? & yes & no\\
  & What is the cat eating? & cake & nothing\\

    \midrule
       \multirow{3}{*}{Irrationality} &        Which bowl has more oranges?	& front & right\\
        & What is the man about to do? &	run & bat\\
        & What is the man doing? & standing & flying kite\\

    \bottomrule
  
\end{tabular}
\caption{Different types of misprediction problems by ViLT \cite{kim2021vilt} when introduced to the visual corruption effects.}
\end{table*}

Understanding this trade-off enables researchers and practitioners to make informed decisions when designing models, striking a balance that aligns with the specific requirements and priorities of the given application. It also highlights the need for comprehensive evaluation metrics that consider both accuracy and robustness, providing a more holistic assessment of model performance. As highlighted in \cref{fig:vreW}, our findings emphasize the delicate interplay between accuracy and robustness in VQA models. Recognizing and managing this trade-off is essential for developing models that align with the desired performance objectives in various real-world scenarios.

\subsection{Mislabeling Problem in Grayscale Images and Color Bias}
\label{sec:mislabeling}
Grayscale images are void of color and hence, the answer to every color-related question on grayscale images should either be unanswerable or a shade of gray. The answers predicted by the model are given full scores as they would match the ground truth color. But grayscaling an image changes the ground truth and hence will require relabeling to prevent inaccurately assessing a model's performance and robustness. As we did not relabel the grayscale images, performance related to grayscale images has not been covered in our work.  

\begin{figure}[htb]
\centering
\includegraphics[width=0.45\textwidth]{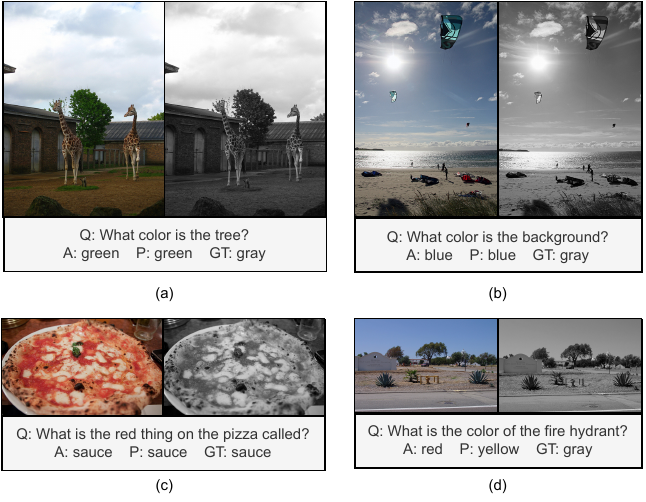}
\caption{Color-based questions on grayscale images causing mislabeling problems for ViLT \cite{kim2021vilt} and indicating the presence of color bias.}
\label{fig:wrong}
\end{figure}

\cref{fig:wrong} highlights a few color-related questions on grayscale images where ViLT \cite{kim2021vilt} predicted a color, indicating that the model associated colors with shapes or structures in the image. As models were able to predict certain colors on images void of color, we can hypothesize that VQA models exhibit some form of color bias. For instance - if the model sees the gray image of an apple, and is asked ``What is the color of the apple?", it will most likely predict \textit{red} since most of the images of apples it was trained on had the color red. Hence, it associated the color red with the shape of the apple. Color bias is caused due to the model's inability to retrieve contextual information from the image as seen in \cite{goyal2017making}.

\subsection{Zero-shot and Robustness}
    \label{sec:zeroShotRobustness}
    Experimental results reveal that the Zero-Shot VQA (ZS-VQA) model PNP \cite{tiong2022plug} is more prone to visual corruption effects compared to traditional methods. However, experiments were conducted on a single ZS-VQA model, the subpar robustness performance cannot be generalized for all ZS-VQA models. PNP exhibits a modular architecture where the overall robustness will depend on the individual robustness of each module. The composing modules: image-question matching module, image captioning module, question-answering module, etc, exhibit different levels of visual robustness. Trivially, we can say that the unimodal question-answering module is unaffected by visual noise while the multimodal image-question matching module and image captioning module are both susceptible to visual noise. 
    
    The low robustness value of PNP can be loosely associated with the low robustness of its composing modules. If the modules are replaced with more robust counterparts, then PNP might become a more robust model. Low robustness scores for ZS-VQA models might seem counter-intuitive as these models are aimed towards handling unseen or out-of-distribution data \cite{teney2016zero, farazi2020known}. By definition, ZS-VQA models should adapt to different contexts and inputs, making them more resilient to variations and uncertainties. This characteristic is particularly valuable in real-world applications where encountering new or unexpected scenarios is common.

\subsection{Unanswerability}
\label{app:unanswerability}

Can the corruption functions render some of the questions unanswerable at certain severity levels? Although this is rare, higher corruption intensities can indeed make some questions unanswerable, even by human evaluators. As the VQA models are classifiers, they should predict the correct answer class as ``unanswerable" or ``nothing". However, similar to grayscale images, this would create a mislabeling problem as the original labels associated with the uncorrupted images were answerable. One plausible solution is manually relabelling the corrupted dataset which can be accurate but costly and impractical.

\section{Additional Future Directions}
\label{app:futureDirection}
 In pursuit of developing a universal robustness evaluation framework, we aim to extend our work by including textual noise, specifically on the input questions. Current literature has explored various forms of textual noise \eg question paraphrasing, semantic error, syntax error \cite{huang2019novel, jimenez2022carets}. Additionally, we propose to simulate typing errors on a physical keyboard \cite{komatsua2018analysis} by associating a probability distribution with each letter being inserted, repeated, removed, replaced, or exchanged with another letter. For instance, the probability of replacement will depend on the proximity of the other letter to the pivot letter based on the layout of the keyboard.

We plan to incorporate consistency metrics \cite{jimenez2022carets} which can be described as an evaluation metric to quantify the model's ability to provide consistent predictions with changes to the input. For instance - for binary classification, if the model predicts 0,1,0,1,0 for five severity levels then it would be deemed inconsistent due to fluctuating predictions. We wish to explore the similarities and differences between consistency and robustness.

Preprocessing the visual or textual input as a form of denoising \cite{motwani2004survey, jain2008natural} might mitigate the performance drop due to corruption effects. For a particular modality, the user can opt to use white-box preprocessing \ie processing the input, given the corruption type, or black-box preprocessing \ie processing the input without any prior knowledge of the corruption type. A VQA model utilizing a denoising module might produce better robustness scores than standard approaches.

VQA models can also be trained on noisy data \eg noise textual labels \cite{zhang2023learning}. The noisy data can include corrupted images and textual noise on both questions and answers. Additional explainable AI techniques in VQA \cite{li2018tell,bendre2021show} can be used to comprehend the processing of visual information by models trained on noisy data during inferences. Training models on grayscale images while retaining the original color labels can help us understand how models perceive shades of grey and whether they associate a specific shade with a particular color.

\end{document}